\documentclass[10pt,journal,compsoc]{IEEEtran}
\usepackage{times}
\usepackage{epsfig}
\usepackage{epstopdf}
\usepackage{graphics}
\usepackage{graphicx}
\usepackage{amsfonts}
\usepackage{amsmath}
\usepackage{amssymb}
\usepackage{booktabs}
\usepackage{mathrsfs}
\usepackage{amsthm}
\usepackage{color}
\usepackage{subfigure}
\usepackage{algorithm}
\usepackage{algpseudocode}
\usepackage{multirow}
\usepackage{pifont}
\usepackage{bm}
\usepackage{bbding}
\usepackage{indentfirst}
\usepackage{bbm}
\usepackage[T1]{fontenc}
\usepackage{url}
\usepackage{booktabs} 
\usepackage{threeparttable}

\ifCLASSOPTIONcompsoc
  \usepackage[nocompress]{cite}
\else
  \usepackage{cite}
\fi

\begin{document}

\title{Comprehensive Instructional Video Analysis: The COIN Dataset and Performance Evaluation}

\author{Yansong Tang, Jiwen~Lu,~\IEEEmembership{Senior Member,~IEEE}, and Jie~Zhou,~\IEEEmembership{Senior Member,~IEEE}\IEEEcompsocitemizethanks{\IEEEcompsocthanksitem Yansong Tang and Jiwen Lu are with the State Key Lab of Intelligent Technologies and Systems, Beijing National Research Center for Information Science and Technology (BNRist), and the Department of Automation, Tsinghua University, Beijing, 100084, China. Email: tys15@mails.tsinghua.edu.cn; lujiwen@tsinghua.edu.cn.
Jie Zhou is with the State Key Lab of Intelligent Technologies and Systems, Beijing National Research Center for Information Science and Technology (BNRist), Department of Automation, Tsinghua University, and the Tsinghua Shenzhen International Graduate School, Tsinghua University, Shenzhen 518055, China. Email: jzhou@tsinghua.edu.cn.}}

\IEEEtitleabstractindextext{%
\begin{abstract}
Thanks to the substantial and explosively inscreased instructional videos on the Internet, novices are able to acquire knowledge for completing various tasks. Over the past decade, growing efforts have been devoted to investigating the problem on instructional video analysis. However, the most existing datasets in this area have limitations in diversity and scale, which makes them far from many real-world applications where more diverse activities occur.
To address this,
we present a large-scale dataset named as ``COIN'' for COmprehensive INstructional video analysis.
Organized with a hierarchical structure, the COIN dataset contains 11,827 videos of 180 tasks in 12 domains (\textit{e.g.,} vehicles, gadgets, etc.) related to our daily life.
With a new developed toolbox,
all the videos are annotated efficiently with a series of step labels and the corresponding temporal boundaries.
In order to provide a benchmark for instructional video analysis,
we evaluate plenty of approaches on the COIN dataset under five different settings.
Furthermore, we exploit two important characteristics (\textit{i.e.,} task-consistency and ordering-dependency) for localizing important steps in instructional videos.
Accordingly, we propose two simple yet effective methods,
which can be easily plugged into conventional proposal-based action detection models.
We believe the introduction of the COIN dataset will promote the future in-depth research on instructional video analysis for the community.
Our dataset, annotation toolbox and source code are available at \url{http://coin-dataset.github.io}.
\end{abstract}

}

\maketitle

\IEEEdisplaynotcompsoctitleabstractindextext

\IEEEpeerreviewmaketitle

\section{Introduction}
\IEEEPARstart
INSTRUCTION, which refers to ``directions about how something should be done or operated''~\cite{OED}, 
enables novices to acquire knowledge from experts to accomplish different tasks.
Over the past decades, learning from instruction has become an important topic in various areas such as cognition science~\cite{koedinger2012knowledge}, educational psychology~\cite{Nadolski2005Optimizing}, and the intersection of computer vision, nature language processing and robotics~\cite{misra2016tell,DBLP:conf/cvpr/AndersonWTB0S0G18,DBLP:journals/pami/AlayracBASLL18}.

Instruction can be expressed by different mediums like text, image and video.
Among them, instructional videos provide more intuitive visual examples, and will be focused on in this paper.
With the explosion of video data on the Internet, people around the world have
uploaded and watched substantial instructional videos~\cite{DBLP:conf/cvpr/AlayracBASLL16,Sener_2015_ICCV}, covering miscellaneous categories.
As suggested by the scientists in educational psychology~\cite{Nadolski2005Optimizing},
novices often face difficulties in learning from the whole realistic task,
and it is necessary to divide the whole task into smaller segments or steps as a form of simplification.
Accordingly, a variety of relative tasks have been studied by morden computer vision community in recent years (\textit{e.g.,} action temporal localization~\cite{DBLP:conf/iccv/ZhaoXWWTL17, DBLP:conf/iccv/XuDS17}, video summarization~\cite{DBLP:journals/pami/ElhamifarSS16,DBLP:conf/eccv/ZhangCSG16,DBLP:journals/tip/PandaMR17} and video caption~\cite{DBLP:journals/corr/abs-1804-00819,DBLP:conf/iccv/KrishnaHRFN17,Yu_2018_CVPR}, etc). 
Also, increasing efforts have been devoted to exploring different challenges of instructional video analysis~\cite{DBLP:conf/cvpr/HuangLFN17,DBLP:conf/aaai/ZhouXC18,Sener_2015_ICCV,DBLP:conf/cvpr/AlayracBASLL16} because of its great research and application value.
As an evidence, Fig. \ref{fig:survey} shows the growing number of publications in the top venues over the recent ten years.

\begin{figure*}[tb]
\includegraphics[width = \linewidth]{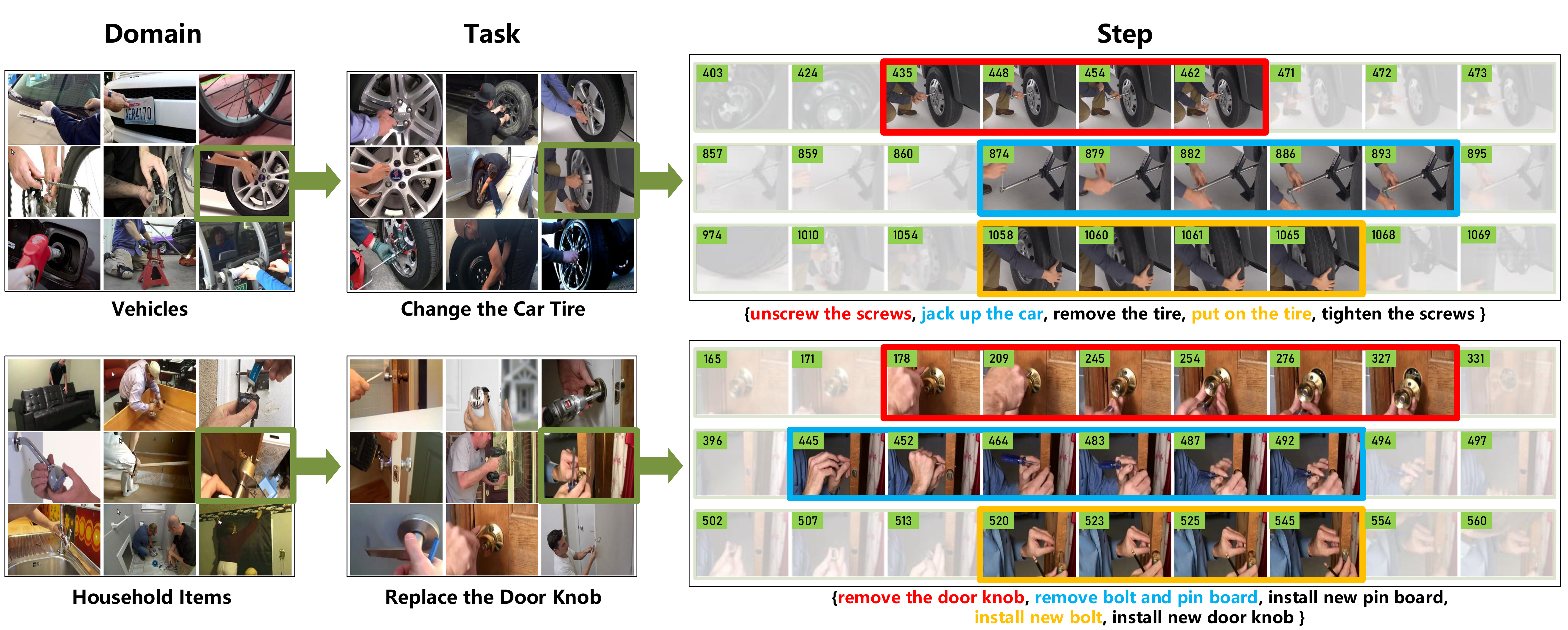}
\caption{Visualization of two root-to-leaf branches of the \textit{COIN}.
There are three levels of our dataset: \textit{domain}, \textit{task} and \textit{step}.
Take the top row as an example, 
in the left box, we show a set of frames of 9 different \textit{tasks} associated with the \textit{domain} ``vehicles''.
In the middle box, we present several images of 9 videos belonging to the \textit{task} ``change the car tire''.
Based on this \textit{task}, in the right box,
we display a sequence of frames sampled from a specific video,
where the indices are presented at the left-top of each frame.
The intervals in red, blue and yellow indicate the \textit{step} of ``unscrew the screws'', ``jack up the car'' and ``put on the tire'',
which are described with the text in corresponding color at the bottom of the right box.
All figures are best viewed in color.
}
\vspace{-0.2cm}
\label{fig:overview}
\end{figure*}

In the meantime, a number of datasets for instructional video analysis~\cite{DBLP:conf/cvpr/RohrbachAAS12, DBLP:conf/cvpr/DasXDC13, DBLP:conf/huc/SteinM13,DBLP:conf/cvpr/KuehneAS14,DBLP:conf/cvpr/AlayracBASLL16,DBLP:conf/dicta/ToyerCHG17,DBLP:conf/aaai/ZhouXC18}
have been collected in the community.
Annotated with texts and temporal boundaries of a series of steps to complete different tasks,
these datasets have provided good benchmarks for preliminary research.
However, most existing datasets focus on a specific domain like cooking, 
which makes them far from many real-world applications where more diverse activities occur.
Moreover, 
the scales of these datasets are insufficient
to satisfy the hunger of recent data-driven learning methods.

To tackle these problems, 
we introduce a new dataset called ``COIN'' for COmprehensive INstructional video analysis.
The COIN dataset contains 11,827 videos of 180 different tasks, 
covering the daily activities related to vehicles, gadgets and many others. 
Unlike the most existing instructional video datasets,
COIN is organized in a three-level semantic structure.
Take the top row of Fig. \ref{fig:overview} as an example,
the first level of this root-to-leaf branch is a \textit{domain} named ``vehicles'',
under which there are numbers of video samples belonging to the second level \textit{tasks}.
For a specific task like ``change the car tire'', 
it is comprised of a series of \textit{steps} such as ``unscrew the screws'', ``jack up the car'', ``put on the tire'', etc.
These \textit{steps} appear in different intervals of a video,
which belong to the third-level tags of COIN.
We also provide the temporal boundaries of all the steps,
which are efficiently annotated based on a new developed toolbox.

Towards the goal to set up a benchmark, 
we implement various approaches on the COIN under five different settings, including step localization, action segmentation, procedure localization, task recognition and step recognition. 
Furthermore,
we propose two simple yet effective methods for localizing different steps in instructional videos.
Specifically, we first explore the \textit{task-consistency} based on bottom-up aggregation and top-down refinement strategies.
Then, we investigate the \textit{ordering-dependency} by considering the transition probability of different steps.
Extensive experimental results have shown the great challenges of COIN and the effectiveness of our methods.
Moreover, we study the cross dataset transfer setting under the conventional ``pre-training+fine-tuning'' paradigm, and demonstrate that COIN can benefit step localization task for other instructional video datasets.

Our main contributions are summarized as follows: 
\begin{itemize}
\item[1)] 
We have introduced the COIN dataset based on our extensive survey on instructional video analysis.
To our best knowledge, this is the currently largest dataset with manual annotation in this field.
Moreover, as a by-product, we have developed an efficient and practical annotation tool, 
which can be further utilized to label temporal boundaries for other tasks like action detection and video caption.
\item[2)] We have evaluated various methods on the COIN dataset under five different evaluation criteria, constructing a benchmark to facilitate the future research. 
Based on the extensive experimental results, we have analyzed our COIN from different aspects, 
and provided an in-depth discussion on the comparison with other relative video analysis datasets.
\item[3)] 
We have exploited the task-consistency and ordering-dependency to further enhance the performance of the step localization task.
Moreover, we have verified that our COIN dataset can contribute to the step localization task for other instructional video datasets.
\end{itemize}

It is to be noted that a preliminary conference version of this work was initially presented in~\cite{COIN}.
As an extension, we have devised a new method to explore the ordering-dependency of different steps in instructional videos, and verified its effectiveness on the COIN and Breakfast\cite{DBLP:conf/cvpr/KuehneAS14} datasets.
Moreover, we have conducted more experiments and provided more in-depth discussions, 
including the study of cross dataset transfer, analysis of hyper-parameters, and a new experimental setting on step recognition.
And also, we have presented a more detailed literature review for instructional video analysis and discussion on future works.

\begin{figure}[tb]
\includegraphics[width = \linewidth]{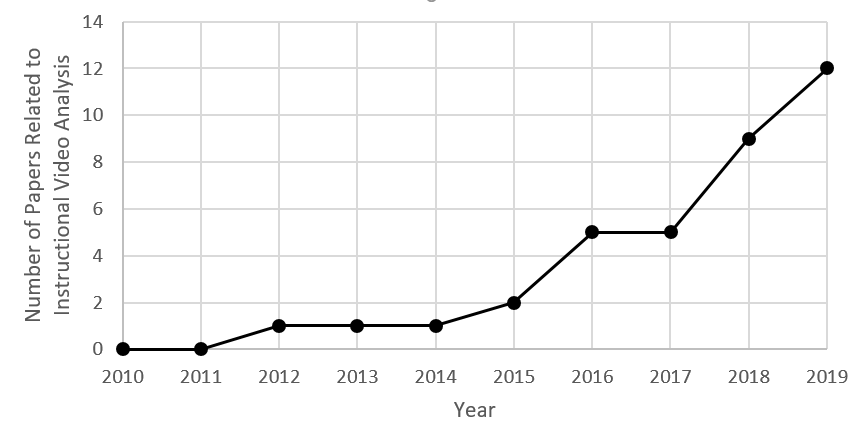}
\caption{
Number of papers related to instructional video analysis published on top computer vision conferences (CVPR/ICCV/ECCV) over the recent 10 years.
}
\vspace{-0.2cm}
\label{fig:survey}
\end{figure}

\linespread{1.2}
\begin{table*}[th]
  \begin{threeparttable}
\caption{\small Comparisons of existing datasets related to instructional video analysis.}
\setlength{\tabcolsep}{8pt}
\begin{tabular}{ r c c c c c c c c c c }
\toprule[1.5pt]
 \textbf{Dataset} & \textbf{Duration} & \textbf{Samples}  & \textbf{Segments} &  \textbf{Task$^*$} & \textbf{Video Source} & \textbf{H?$^{**}$}  & \textbf{Manual Anno.$^{***}$}  & \textbf{Classes} & \textbf{Year}\\
 \midrule
  MPII\cite{DBLP:conf/cvpr/RohrbachAAS12} & 9h,48m & 44 & 5,609 & CA & self-collected &  \ding{55}& SL + TB  & - & 2012 \\ 
  YouCook\cite{DBLP:conf/cvpr/DasXDC13}     & 2h,20m & 88 & - & CA &  YouTube &\ding{55} & SL + TB & - & 2013 \\
  50Salads\cite{DBLP:conf/huc/SteinM13} & 5h,20m & 50 & 966 & CA &  self-collected &  \ding{55}   & SL + TB & - & 2013 \\
  Breakfast\cite{DBLP:conf/cvpr/KuehneAS14} & 77h & 1,989 & 8,456 & CA &  self-collected & \ding{55} & SD + TB &  10 & 2014 \\
  JIGSAWS\cite{gao2014jhu} & 2.6h & 206 & 1,703 & SA &  self-collected & \ding{55} & SD + TB &  3 & 2014 \\  
  ``5 tasks''\cite{DBLP:conf/cvpr/AlayracBASLL16} & 5h & 150 & - & CT &  YouTube & \ding{55}  &  SD + TB &  5 & 2016  \\
  Ikea-FA\cite{DBLP:conf/dicta/ToyerCHG17} & 3h,50m & 101 & 1,911 & AF & self-collected & \ding{55} &  SL + TB & - & 2017  \\
  Recipe1M$^{\dag}$\cite{DBLP:conf/cvpr/SalvadorHAMOW017}   & - & 432 & - & CA & - & \ding{55}  & SD & - & 2017  \\    
  Recipe1M+$^{\dag}$\cite{DBLP:journals/tpami/recipe1M}   & - & 13,735,679 & - & CA & Google & \ding{55}  & SD & - & 2018  \\      
  YouCook2\cite{DBLP:conf/aaai/ZhouXC18} & 176h & 2,000 & 13,829 & CA & YouTube & \ding{55}  & SD + TB + OBL & 89 & 2018  \\
  EPIC-KITCHENS\cite{Damen_2018_ECCV}   & 55h & 432 & 39,596 & CA & self-collected & \ding{55} & SD + TB + OBL &5 & 2018  \\  
  EPIC-Skills\cite{EPIC-skill}   & 5.2h & 216 & - & - & mixed$^{\dag\dag}$ & \ding{55}  & WB & 4 & 2018  \\        
  CrossTask\cite{cross-task}   & 376h & 4,700 & - & CT & YouTube & \ding{51} & SL + TB & 83  & 2019  \\    
  BEST\cite{BEST}   & 26h & 500 & - & CT & YouTube & \ding{55} & WB  & 5 & 2019  \\ 
  HowTo100M\cite{miech19howto100m}   & 134,472h & 1.22M & 136M & CT & YouTube & \ding{51} & -  &  23K & 2019  \\  
\midrule
  \textbf{COIN} (Ours) & 476h,38m & 11,827 & 46,354 & CT & YouTube & \ding{51} & SL + TB & 180 & \\  
\bottomrule[1.5pt]
\end{tabular}
\label{tab:tabstata}
\begin{tablenotes}
  \footnotesize
  \item[$\dag$] Both Recipe1M and Recipe1M+ are image-based datasets.
  \quad $\dag\dag$ The EPIC-Skills dataset comprised of four tasks, where two were self-recorded, and two were seletcted from published datasets~\cite{gao2014jhu, de2009guide}.
  \quad $^{*}$ CA: cooking activities,  SA: surgical activities, AF: assembling furniture, CT: comprehensive tasks  \\  $^{**}$ H?: Hierarchical? \quad
  $^{***}$ SL: step label (shared in different videos), SD: step description (not shared in different videos), TB: temporal boundary, \\ OBL: object bounding box and label, WB: which is better given a pair-wise videos.  
\end{tablenotes}
\end{threeparttable}
\vspace{-0.2cm}
\end{table*}

\linespread{1}

\section{Related Work}
\subsection{Tasks for Instructional Video Analysis}
\begin{figure}[tb]
\includegraphics[width = \linewidth]{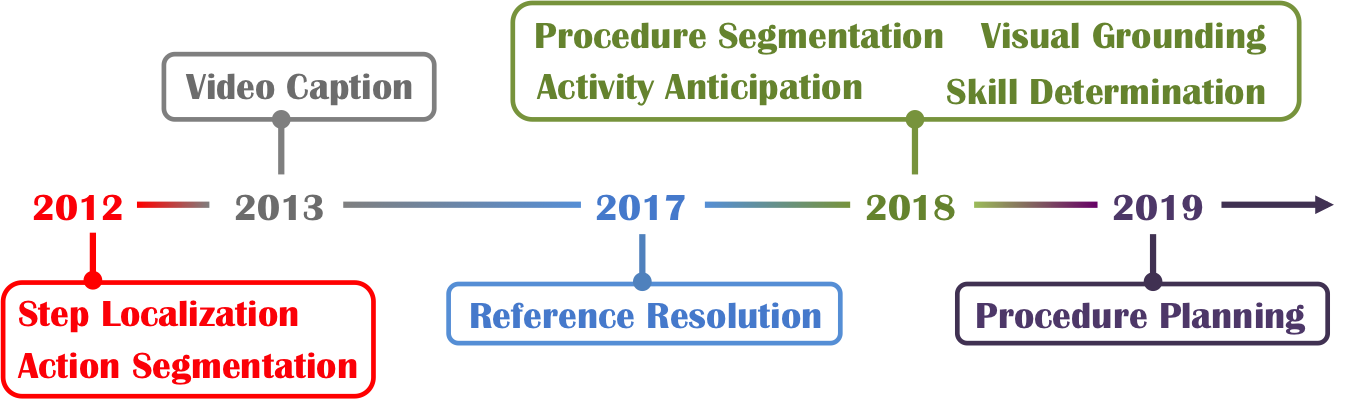}
\caption{The timeline of different tasks for instructional video analysis being proposed.
}
\vspace{-0.2cm}
\label{fig:history}
\end{figure}

Fig. \ref{fig:history} shows the development of different tasks for instructional video analysis being proposed in past decade.
In 2012, Rohrbach \textit{et al.}~\cite{DBLP:conf/cvpr/RohrbachAAS12} collected the MPII dataset, 
which promoted the later works on \textit{step localization} and \textit{action segmentation}.
As the two fundamental tasks in this field, 
\textit{step localization} aims to  localize the start and end points of a series of steps and recognize their labels, 
and \textit{action segmentation} targets to parse a video into different actions at frame-level.
In later sections of the paper we will concentrate more on these two tasks,
while in this subsection we provide some brief introduction on other tasks.

In 2013, 
Das \textit{et al.}\cite{DBLP:conf/cvpr/KuehneAS14} proposed the YouCook dataset and facilitated the research on video caption,
which required generating sentences to describe the videos.
In 2017, Huang \textit{et al.}~\cite{DBLP:conf/cvpr/HuangLFN17} studied the task of reference resolution, which aimed to \textit{temporally}
link an entity to the original action that produced it.
In 2018, they further investigated the visual grounding problem~\cite{Huang_2018_CVPR}, 
which explored the visual-linguistic meaning of referring expressions in both \textit{spatial} and \textit{temporal} domains.
In the same year, Zhou \textit{et al.}~\cite{DBLP:conf/aaai/ZhouXC18} presented a procedure segmentation task, 
targeting at segmenting an instructional video into category-independent procedure segments.
Farha \textit{et al.}~\cite{DBLP:conf/cvpr/FarhaRG18} studied the activity anticipation problem,
which predicted the future actions and their durations in instructional videos.
Doughty \textit{et al.}~\cite{EPIC-skill} addressed the issue on skill determination, 
which assessed the skill behaviour of a subject.
More recently, Chang \textit{et al.}~\cite{DBLP:journals/corr/abs-1907-01172} presented a new task of procedure planning in instructional videos,
which aimed to discover the intermediate actions according to the start and final observations.
With these promotion of the new topics on instructional video analysis, 
the research community is paying growing attention to this burgeoning field.

\subsection{Datasets Related to Instructional Video Analysis}
There are mainly three types of related datasets.
(1) The action detection datasets are comprised of untrimmed video samples,
and the goal is to recognize and localize the action instances on temporal domain~\cite{DBLP:conf/cvpr/HeilbronEGN15, THUMOS14,DBLP:journals/tip/TangLWYZ19} or spatial-temporal domain~\cite{Gu_2018_CVPR}.
(2) The video summarization datasets~\cite{DBLP:journals/prl/AvilaLLA11,DBLP:conf/eccv/GygliGRG14,DBLP:conf/cvpr/SongVSJ15,DBLP:journals/tip/PandaMR17} contain long videos arranging from different domains. The objective is to extract a set of informative frames in order to briefly summarize the video content.
(3) The video caption datasets are annotated with descried sentences or phrases,
which can be based on either a trimmed video~\cite{DBLP:conf/cvpr/XuMYR16, Yu_2018_CVPR} or different segments of a long video~\cite{DBLP:conf/iccv/KrishnaHRFN17}.
Our COIN is relevant to the above mentioned datasets, 
as it requires to localize the temporal boundaries of important steps corresponding to a task.
The main differences lie in the following two aspects:
(1) \textbf{Task-consistency.} 
The steps belonging to different tasks shall not appear in the same video.
For example, it is unlikely for an instructional video to contain the step ``pour water to the tree'' (belongs to task ``plant tree'') and the step ``install the lampshade'' (belongs to task ``replace a bulb'').
(2) \textbf{Ordering-dependency.}
There may be some intrinsic ordering constraints among a series of steps for completing different tasks. 
For example, for the task of ``planting tree'',
the step ``dig a hole'' shall be ahead of the step ``put the tree into the hole''.

There have been a variety of instructional video datasets proposed in recent years,
and we briefly review some representative datasets in supplementary.
Table \ref{tab:tabstata} summarizes the comparison among some publicly relevant instructional datasets and our proposed COIN.
While the existing datasets present various challenges to some extent, they still have some limitations in the following two aspects.
(1) \textbf{Diversity:} Most of these datasets tend to be specific and contain certain types of instructional activities, \textit{e.g.,} cooking.
However, according to some widely-used websites~\cite{howcast, howdini, wikihow}, people attempt to acquire knowledge from various types of instructional video across different domains.
(2) \textbf{Scale:}
Compared with the recent datasets for image classification (e.g., ImageNet~\cite{DBLP:conf/cvpr/DengDSLL009} with ~1 million images) and action detection (e.g., ActivityNet v1.3~\cite{DBLP:conf/cvpr/HeilbronEGN15} with ~20k videos),
most existing instructional video datasets are relatively smaller in scale.
Though the HowTo100M dataset provided a great amount of data, its automaticly generated annotation might be inaccurate as the authors mentioned in~\cite{miech19howto100m}.
The challenge of building such a large-scale dataset mainly stems from the difficulty to organize enormous amount of video and the heavy workload of annotation. 
To address these two issues, 
we first establish a rich semantic taxonomy covering 12 domains and collect 11,827 instructional videos to construct COIN. 
With our new developed toolbox,
we also provide the temporal boundaries of steps that appear in all the videos with efficient and precise annotation.

\subsection{Methods for Instructional Video Analysis}
In this subsection, we review a series of approaches related to two core tasks for instructional video analysis (step localization and action segmentation).
We roughly divided them into three categories according to experimental settings: unsupervised learning-based, weakly-supervised learning-based and fully-supervised learning-based.

\textbf{Unsupervised Learning Approaches:}
In the first category,
the step localization task usually took a video and the corresponding narration or subtitle as multi-modal inputs~\footnote{The language signal should not be treated as supervision since the steps are not directly given, but need to be further explored in an unsupervised manner.}.
For example,
Sener \textit{et al.}\cite{Sener_2015_ICCV} developed a joint generative model to parse both video frames and subtitles into activity steps.
Alayrac \textit{et al.}\cite{DBLP:conf/cvpr/AlayracBASLL16} leveraged the complementary nature of the instructional video and its narration
to discover and locate the main steps of a certain task.
Generally speaking, the advantages of employing the narration or subtitle is to avoid human annotation, which may cost huge workload.
However, these narration or subtitles may be inaccurate~\cite{DBLP:conf/aaai/ZhouXC18} or even irrelevant to the video as we mention above.
For the action segmentation task,
Aakur \textit{et al.}~\cite{Aakur_2019_CVPR} presented a self-supervised and predictive learning framework to explore the spatial-temporal dynamics of the videos,
while Sener \textit{et al.}~\cite{Sener_2018_CVPR} proposed a Generalized Mallows Model (GMM) to model the distribution over sub-activity permutations.
More recently,
Kukleva \textit{et al.}~\cite{Kukleva_2019_CVPR} first learned a continuous temporal embedding of frame-based features, 
and then decoded the videos into coherent action segments according to an ordered clustering of these features.

\textbf{Weakly-supervised Learning Approaches:}
In the second category,
the step localization problem has recently been studied by Zhukov \textit{et al.}~\cite{cross-task} by exploring the sharing information of different steps across different tasks.
And Liu \textit{et al.}~\cite{DBLP:conf/cvpr/LiuJ019} 
identified and addressed the action completeness
modeling and action-context separation problems for temporal action localization by the weak supervision.
For the action segmentation task,
Kuehne \textit{et al.}~\cite{DBLP:conf/cvpr/KuehneAS14} developed a hierarchical model based on HMMs and a context-free grammar to parse the main steps in the cooking activities. Richard \textit{et al.}~\cite{richard2017action}\cite{DBLP:journals/corr/abs-1805-06875} adopted Viterbi algorithm to solve the probabilistic model of weakly supervised segmentation. Ding \textit{et al.}~\cite{DBLP:journals/corr/abs-1803-10699} proposed a temporal convolutional feature pyramid network to predict frame-wise labels and use soft boundary assignment to iteratively optimize the segmentation results.
In this work, we also evaluate these three methods\footnote{The details of the weak supervisions are described in section 5.2.} to provide some baseline results on COIN.
More recently,
Chang \textit{et al.}~\cite{Chang_2019_CVPR} developed a discriminative differentiable dynamic time warping (D$^3$TW) method,
which extended the ordering loss to be differentiable.

\textbf{Fully-supervised Learning Approaches:}
In the third category, 
the action segmentation task has been explored by numbers of works by developing various types of network architectures. 
For example, multi-stream bi-directional recurrent neural network (MSB-RNN)~\cite{DBLP:conf/cvpr/SinghMJTS16}, temporal deformable residual network (TDRN)~\cite{DBLP:conf/cvpr/LeiT18}, multi-stage temporal convolutional network (MS-TCN)~\cite{Farha_2019_CVPR}, etc.
As a task we pay more attention to,
the step localization is related to the area of action detection,
where promising progress has also been achieved recently~\cite{DBLP:conf/cvpr/Liu0Z0C19, DBLP:conf/cvpr/LongYQTLM19,BSN,BMN}.
For example,
Zhao \textit{et al.}~\cite{DBLP:conf/iccv/ZhaoXWWTL17} developed structured segment networks (SSN) to model the temporal structure of each action instance with a structured temporal pyramid.
Xu \textit{et al.}~\cite{DBLP:conf/iccv/XuDS17} introduced a Region Convolutional 3D Network (R-C3D) architecture,
which was built on C3D~\cite{DBLP:conf/iccv/TranBFTP15} and Faster R-CNN~\cite{faster-rcnn},
to explore the region information of video frames.
Compared with these methods,
we attempt to further explore the dependencies of different steps, which lies in the intrinsic structure of instructional videos.
Towards this goal, we proposed two methods to leverage the task-consistency and ordering-dependency of different steps.
Our methods can be easily plugged into recent proposal-based action detection methods and enhance the performance of step localization task for instructional video analysis.

\section{The COIN Dataset}
\begin{figure}[tb]
\includegraphics[width = \linewidth]{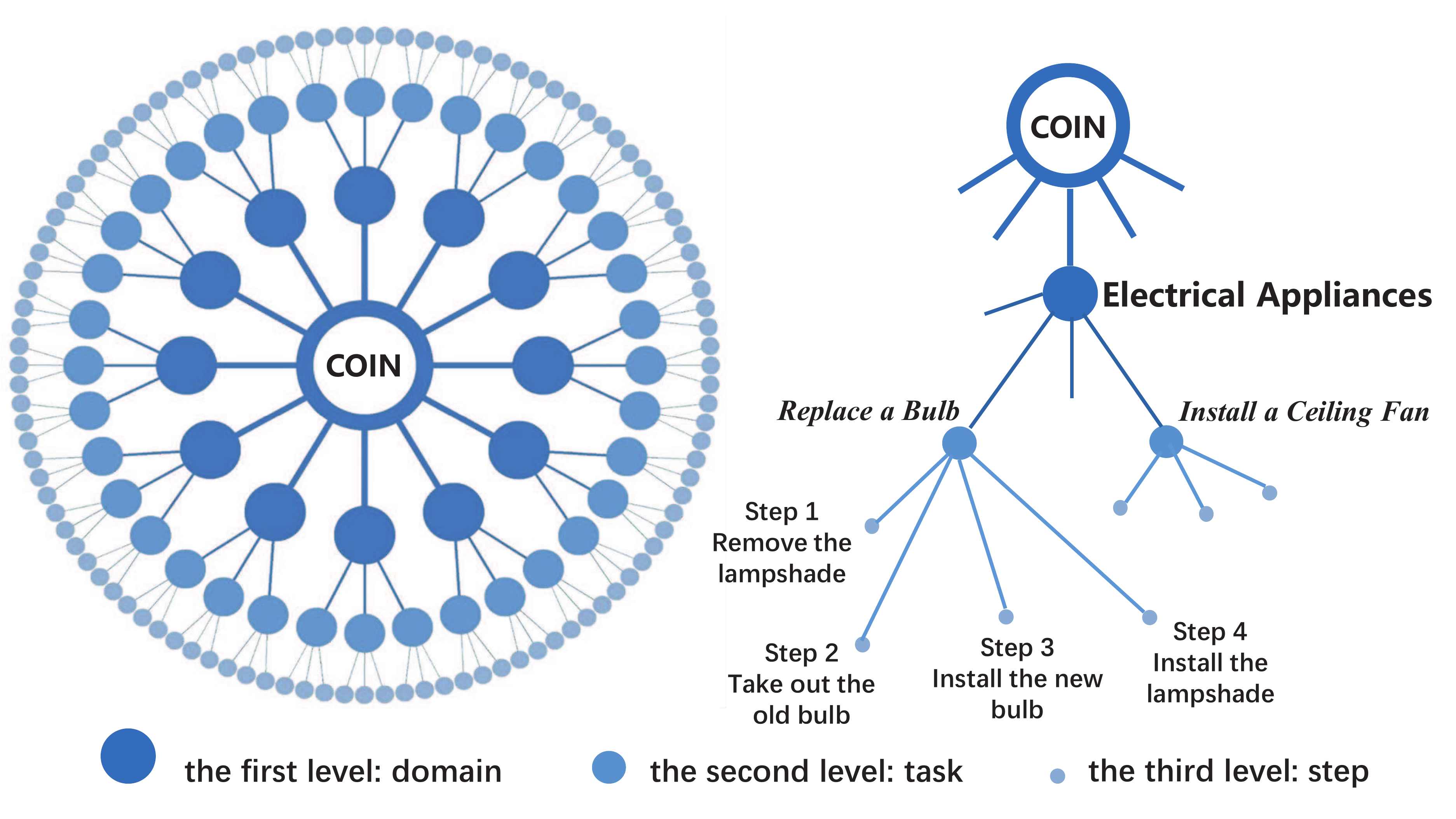}
\caption{Illustration of the COIN lexicon.
The left figure shows the hierarchical structure, 
where the nodes of three different sizes correspond to the domain, task and step respectively. For brevity, we do not draw all the tasks and steps here.
The right figure presents detailed steps of the task ``replace a bulb",
which belongs to the domain ``electrical appliances".
}
\label{fig:tag}
\end{figure}

\subsection{Lexicon}
The purpose of COIN is to establish a rich semantic taxonomy to organize comprehensive instructional videos.
In previous literature,
some representative large-scale datasets were built upon existing structures.
For example, 
the ImageNet~\cite{DBLP:conf/cvpr/DengDSLL009} database was constructed based on a hierarchical structure of WordNet~\cite{WordNet},
while the ActivityNet dataset~\cite{DBLP:conf/cvpr/HeilbronEGN15} adopted the activity taxonomy organized by American Time Use Survey (ATUS)~\cite{ATUS}.
In comparison, 
it remains great difficulty to define such a semantic lexicon for instructional videos because of their high diversity and complex temporal structure.
Hence,
most existing instructional video datasets~\cite{DBLP:conf/aaai/ZhouXC18} focus on a specific domain like cooking or furniture assembling, 
and ~\cite{DBLP:conf/cvpr/AlayracBASLL16} only consists of five tasks.
Towards the goal to construct a large-scale benchmark with high diversity, 
we utilize a hierarchical structure to organize our dataset.
Fig. \ref{fig:overview} and Fig. \ref{fig:tag} present the illustration of our lexicon,
which contains three levels from roots to leaves: domain, task and step.

(1) \textbf{Domain:}
For the first level,
we bring the ideas from the organization of several websites\cite{howcast}\cite{wikihow}\cite{howdini},
which are commonly-used for users to watch or upload instructional videos.
We choose 12 domains as: \textit{nursing \& caring}, \textit{vehicles}, \textit{leisure \& performance}, \textit{gadgets}, \textit{electric appliances}, \textit{household items}, \textit{science \& craft}, \textit{plants \& fruits}, \textit{snacks \& drinks}, \textit{dishes}, \textit{sports}, and \textit{housework}.

(2) \textbf{Task:}
As the second level, the task is linked to the domain.
For example, the tasks ``replace a bulb'' and ``install a ceiling fan'' are associated with the domain ``electrical appliances''.
As most tasks on \cite{howcast}\cite{wikihow}\cite{howdini} may be too specific,
we further search different tasks of the 12 domains on YouTube.
In order to ensure the tasks of COIN are commonly used,
we finally select 180 tasks, under which the searched videos are often viewed~\footnote{We present the statistics of browse times in supplementary material.}.

(3) \textbf{Step:}
The third level of the lexicon contains various series of steps to complete different tasks.
For example,
steps ``remove the lampshade'', ``take out the old bulb'', ``install the new bulb'' and ``install the lampshade'' are associated with the tasks ``replace a bulb''.
We employed 6 experts (\textit{e.g.,} driver, athlete, etc.) who have prior knowledge in the 12 domains to define these steps.
They were asked to browse the corresponding videos as a preparation in order to provide the high-quality definition,
and each step phrase will be double checked by another expert.
In total, there are 778 defined steps.
Note that we do not directly adopt narrated information,
which might have large variance for a specific task,
because we expect to obtain the simplification of the core steps,
which are common in different videos of accomplishing a certain task.

\subsection{Annotation Tool}
\begin{figure}[tb]
\setlength{\abovecaptionskip}{-0.1cm}
\includegraphics[width = \linewidth]{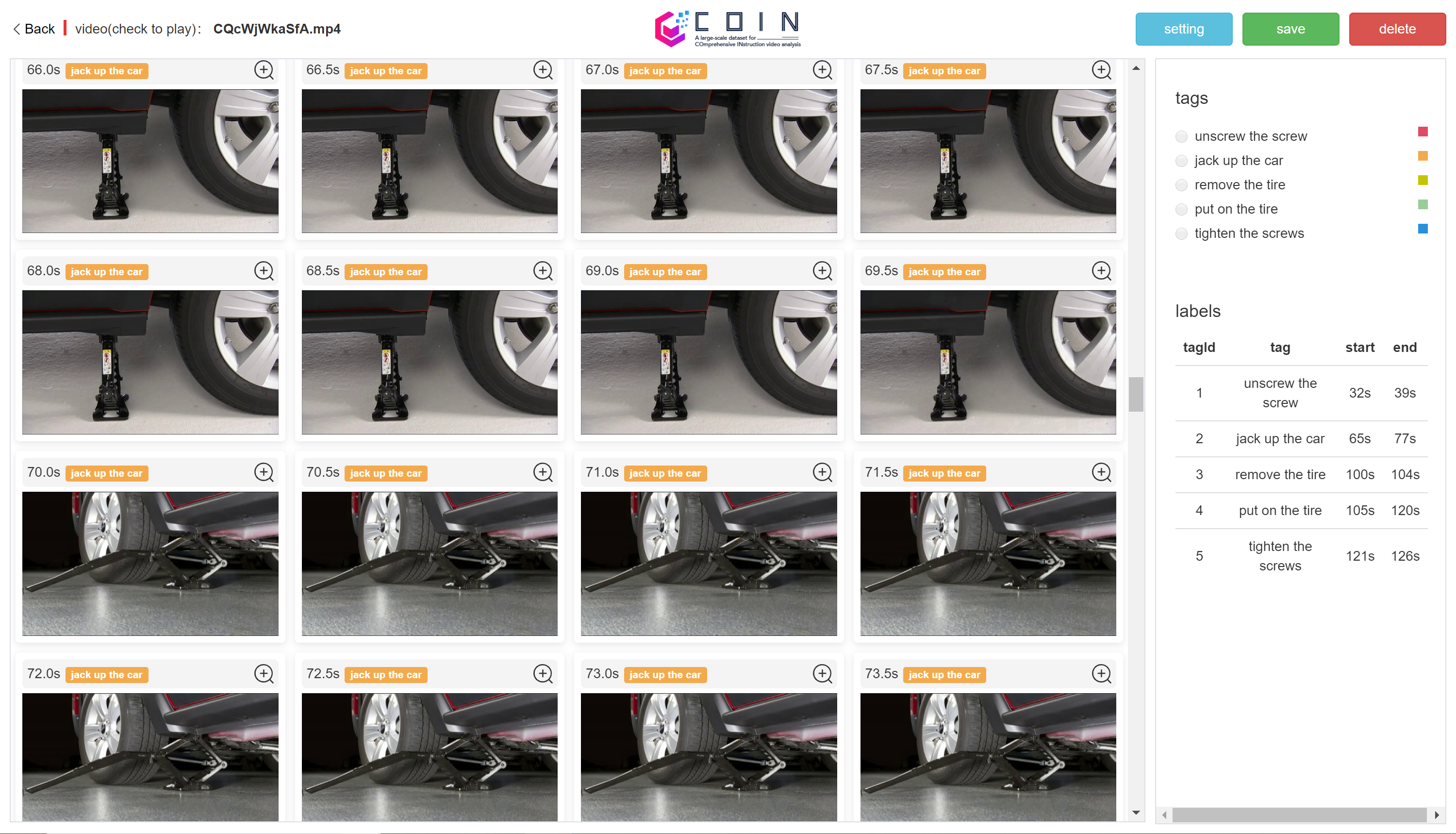}
\caption{The interface of our new developed annotation tool under the \textit{frame mode}.}
\label{fig:tool}
\end{figure}

Given an instructional video,
the goal of annotation is to label the step categories and the corresponding segments.
As the segments are variant in length and content, 
it will cost huge workload to label the COIN with conventional annotation tool.
In order to improve the annotation efficiency, we develop a new toolbox which has two modes: \textit{frame mode} and \textit{video mode}.
Fig. \ref{fig:tool} shows an example interface of the \textit{frame mode}, which presents the frames extracted from a video under an adjustable frame rate (default is 2fps).
Under the \textit{frame mode}, the annotator can directly select the start and end frame of the segment as well as its label. 
However, 
due to the time gap between two adjacent frames,
some quick and consecutive actions might be missed.
To address this problem,
we adopt another \textit{video mode}.
The \textit{video mode} of the annotation tool presents the online video and timeline, which is frequently used in previous video annotation systems~\cite{DBLP:conf/iccv/KrishnaHRFN17}. 
Though the \textit{video mode} brings more continuous information in the time scale, 
it is much more time-consuming than the \textit{frame mode} because of the process to locate a certain frame and adjust the timeline\footnote{
For a set of videos, the annotation time under the \textit{frame mode} is only 26.8\% of that under the \textit{video mode}. Please see supplementary material for details.}.

During the annotation process,
each video is labelled by three different workers with payments.
To begin with, the first worker generated primary annotation under the \textit{frame mode}. 
Next, the second worker adjusted the annotation based on the results of the first worker. 
Ultimately, the third worker switched to the \textit{video mode} to check and refine the annotation.
Under this pipeline,
the total time of the annotation process is about 600 hours.

\begin{figure}[tb]
\includegraphics[width = \linewidth]{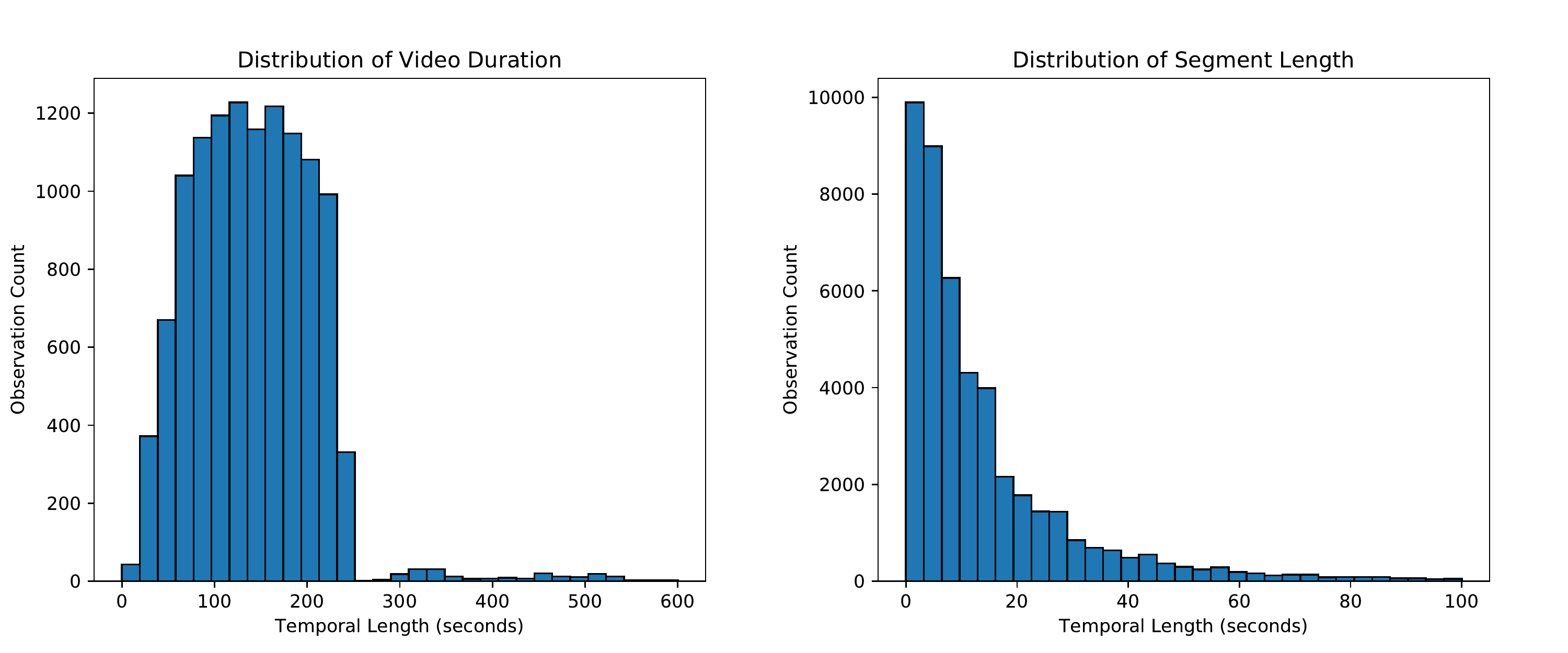}
\caption{
The duration statistics of the videos (left) and segments (right) in the COIN dataset.
}
\vspace{-0.2cm}
\label{fig:dur_sta}
\end{figure}

\subsection{Statistics}
The COIN dataset consists of 11,827 videos related to 180 different tasks, 
which were all collected from YouTube. 
We split the COIN into 9030 and 2797 video samples for training and testing respectively, 
and show the sample distributions among all the task categories in supplementary.
Fig. \ref{fig:dur_sta} displays the duration distribution of videos and segments.
The averaged length of a video is 2.36 minutes.
Each video is labelled with 3.91 step segments, where each segment lasts 14.91 seconds on average.
In total, the dataset contains videos of 476 hours, with 46,354 annotated segments.

In order to further illustrate different charaterstics of the COIN dataset, we calculate two scores which are similarly used in~\cite{DBLP:journals/pami/AlayracBASLL18}. 
For a given task, suppose there are N video samples and K steps defined in the ground truth,
for the $n-th$ video, 
let $u^{(n)}$ denote the numbers of the unique anotated steps,
$l^{(n)}$ denote the length of the longest common subsequence between the annotated
sequence of steps and the ground truth sequence,
and $v^{(n)}$ denote the number of annotated steps without duplicate.
Then the missing steps score (MSS) and order consistency error (OCE) are defined as below:
\begin{equation}
  \text{MSS} := 1 - \dfrac{\sum\limits_{n=1}^N v^{(n)}}{K \cdot N}, \quad
  \text{OCE} := 1 - \dfrac{\sum\limits_{n=1}^N l^{(n)}}{\sum\limits_{n=1}^N u^{(n)}}.
  \label{eq:sta}
\end{equation}

The average values of the 180 tasks in the COIN are 0.2924 (MSS) and 0.2076 (OCE) respectively, 
and the concrete values of each task are presented in the supplementary.
The statistics illustrates that videos of the same task would not strictly share the same series of ordered steps due to the abbreaviated or reversed step sequences.
For example, a task ``A'' might contains different step sequences of $\{a_1, a_2, a_3\}$, $\{a_1, a_2, a_3, a_4\}$, $\{a_1, a_3, a_2\}$.

\section{Approach}

\subsection{Task-consistency Analysis}
\begin{figure}[tb]
\includegraphics[width = \linewidth]{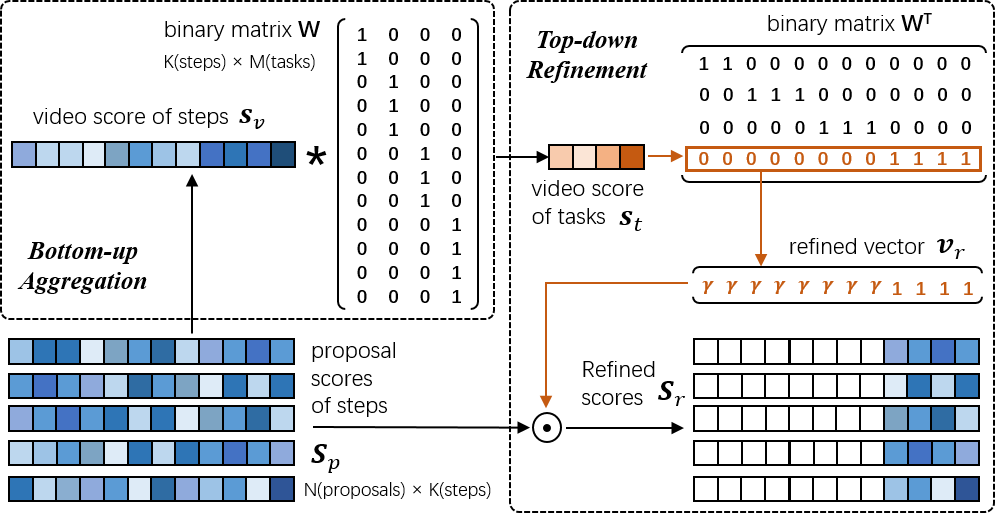}
\caption{Flowchart of our proposed task-consistency method.
During the first \textbf{bottom-up aggregation} stage,
the inputs are a series of scores $\textbf{S}_p=\{\textbf{s}_p^1, ... , \textbf{s}_p^n, ..., \textbf{s}_p^N\}$ of an instructional video,
which denotes the probabilities of each step appearing in the corresponding proposal.
We first aggregate them into a video-based score $\textbf{s}_v$, 
and map it into another score $\textbf{s}_t$ to predict the task label $Y$.
At \textbf{top-down refinement} stage,
we generate a refined mask vector $\textbf{v}_r$ based on the task label.
Then we alleviate the weights of other bits in $\textbf{S}_p$ by $\textbf{v}_r$ to ensure the task-consistency.
The refined scores $\textbf{S}_r$ are finally utilized to perform NMS process and output the final results.
}
\label{fig:domain}
\end{figure}

Given an instructional video, 
one important real-world application is to localize a series of steps to complete the corresponding task.
In this section, we introduce a new proposed task-consistency method for step localization in instructional videos.
Our method is motivated by the intrinsic dependencies of different steps which are associated to a certain task.
For example,
it is unlikely for the steps of ``dig a pit of proper size'' and ``soak the strips into water''
to occur in the same video, because they belong to different tasks of ``plant tree'' and ``make french fries'' respectively.
In another word,
the steps in the same video should be task-consistent to ensure that they belong to the same task.
Fig. \ref{fig:domain} presents the flowchart of our task-consistency method,
which contains two stages: (1) bottom-up aggregation and (2) top-down refinement.

\textbf{Bottom-up Aggregation:}
As our method is built upon the proposal-based action detection methods,
we start with training an existing action detector, \textit{e.g.,} SSN~\cite{DBLP:conf/iccv/ZhaoXWWTL17}, on our COIN dataset.
During inference phase, given an input video, we send it into the action detector to produce a series of proposals with their corresponding locations and predicted scores.
These scores indicate the probabilities of each step occuring in the corresponding proposal.
We denote them as $\textbf{S}_p = \{\textbf{s}_p^1, ..., \textbf{s}_p^n, ..., \textbf{s}_p^N\}$, 
where $\textbf{s}_p^n \in R^K$ represents the score of the $n-th$ proposal and $K$ is the number of the total steps.
The goal of the bottom-up aggregation stage is to predict the task labels based on these proposal scores.
To this end, we first aggregate the scores along all the proposals as $\textbf{s}_v = \sum_{n=1}^N \textbf{s}_p^n$,
where $\textbf{s}_v$ indicates the probability of each step appearing in the video.
Then we construct a binary matrix $W$ with the size of $K \times M$
to model the relationship between the $K$ steps and $M$ tasks: 
\begin{eqnarray}
    w_{ij} = 
    \begin{cases}
1, \quad \text{if step $i$ belongs to task $j$}\\
0, \quad \text{otherwise} \\
    \end{cases}
\end{eqnarray}

Having obtained the step-based score $\textbf{s}_v$ and the binary matrix $W$,
we calculate a task-based score as $\textbf{s}_t = \textbf{s}_v W$. 
This operation is essential to combine the scores of steps belonging to same tasks. 
We choose the index $Y$ with the max value in the $\textbf{s}_t$ as the task label of the entire video.

Note that since we sum the step scores to calculate the task score,
someone might cast doubt that the task score will be overwhelmed by the task with more steps.
Actually, besides summing the scores, another way is averaging them based on the number of steps.
In fact, these two methods equally weigh all the steps or tasks respectively.
We conduct experiments in section 5 and find the summing method is more effective on the COIN.

\textbf{Top-down Refinement:}
The target of the top-down refinement stage is to refine the original proposal scores with the guidance of the task label.
We first select the $Y-th$ row in $W$ as a mask vector $\textbf{v}$,
based on which we define a refined vector as:
\begin{eqnarray}
 \textbf{v}_r = \textbf{v} + \gamma (\textbf{I} - \textbf{v}) .
\end{eqnarray}

Here $\textbf{I}$ is an vector where all the elements equal to 1.
$\gamma$ is an attenuation coefficient to alleviate the weights of the steps which do not belong to the task $Y$.
We empirically set $\gamma$ to be $e^{-2}$, and 
further exploration on this parameter can be found in our supplementary.
Then, we employ the $\textbf{v}_r$ to mask the original scores $\textbf{s}_p^n$ as:
\begin{equation}
    \textbf{s}_r^n = \textbf{s}_p^n \odot \textbf{v}_r ,
\end{equation}
where $\odot$ is the element-wise Hadamard product.
We compute a sequence of scores as $\textbf{S}_r = \{\textbf{s}_r^{1}, ..., \textbf{s}_r^{n}, ..., \textbf{s}_r^{N}\}$.
Based on these refined scores and their locations,
we employ a Non-Maximum Suppression (NMS) strategy to obtain the results of step localization.
In summary, we first predict the task label through the bottom-up scheme,
and refine the proposal scores by the top-down strategy,
hence the task-consistency is guaranteed.

\begin{figure}[tb]
\includegraphics[width = \linewidth]{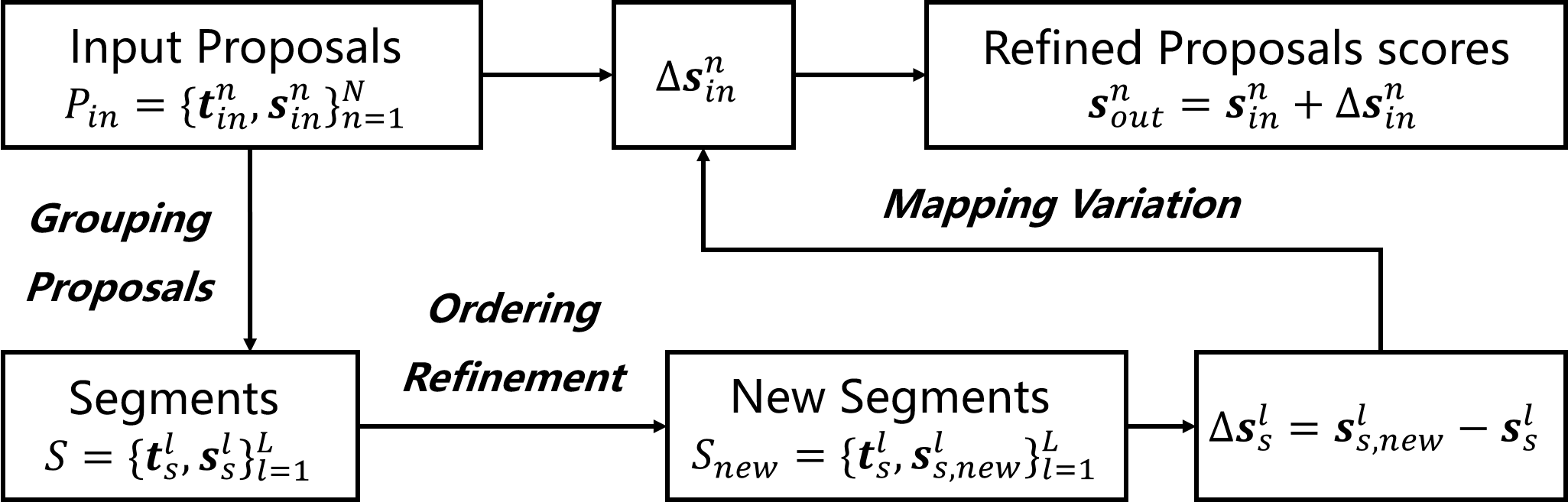}
\caption{Pipeline of our ordering-dependency method.
The input of our method is a set of proposals $P_{in}$ and we refine it through the following three stages.
(1) In order to deal with the overlapped proposals $P_{in}$, we first group them into a series of segments $S$.
(2) We refine the segments $S$ based on the ordering information in the training set.
(3) We map the score variation of $\Delta \textbf{s}_s^l$ into $\Delta \textbf{s}_{in}^n$, and finally refine the proposal $P_{in}$.
}
\label{fig:ordering}
\end{figure}

\subsection{Ordering-dependency Analysis}
Another important characteristic of instructional video is the ordering constraint of different steps belonging to a task.
Some previous works~\cite{DBLP:journals/pami/AlayracBASLL18,DBLP:conf/cvpr/AlayracBASLL16} assumed that 
videos of the same task share the same series of ordered steps.
However this assumption is too strict as several steps might be abbreviated or reversed based on the statistics we show in Section 3.3.
To address this issue, we propose a new method to perform ordering refinement by leveraging the transition probability between different steps.
Fig. \ref{fig:ordering} displays a pipeline of our approach which consists of three stages as
(1) grouping proposals, (2) ordering refinement and (3) mapping variation.
We elaborate each stage in detail as follow.

\textbf{Grouping Proposals $P_{in}$ into Segment $S$.}
The ``grouping proposals'' can be considered as a transformation from a proposal space $\mathbb{P}$ to a segment space $\mathbb{S}$.
Similar to the task-consistency method, 
the input of our ordering-dependency approach is a set of proposals $P_{in}=\{\textbf{t}_{in}^n,\textbf{s}_{in}^n\}_{n=1}^N \in \mathbb{P}$ which are generated by existing proposal-based action detector.
Here $\textbf{t}_{in} = [t^{n1}_{in},t^{n2}_{in}]$ denotes its temporal interval, and $\textbf{s}_{in}^n$ is the corresponding step probability scores need to be refined. 
However, these detectors usually produce a great amount of proposals with many overlaps, which make the ordering refinement hard to perform.
This is because (1) the overlaps might bring ambiguity for deciding which proposal occurs earlier, and (2) several overlapped proposals might actually correspond to the same step.
To address this, we group the proposals into a sequence of segments $S = \{\textbf{t}^l_s,\textbf{s}^l_s\}_{l=1}^L\in \mathbb{S}$ where the overlaps are eliminated.
Specifically, for an input proposal $P_{in}^n$, we first generate a Gaussian based function as below:
\begin{eqnarray} 
\textbf{f}^n(t) = \frac{\textbf{s}^n_{in}}{\sqrt{2\pi}\sigma_n}e^{-\frac{(t-\mu_n)^2}{2\sigma_n^2}}. 
\end{eqnarray}

\begin{figure}[tb]
\includegraphics[width = \linewidth]{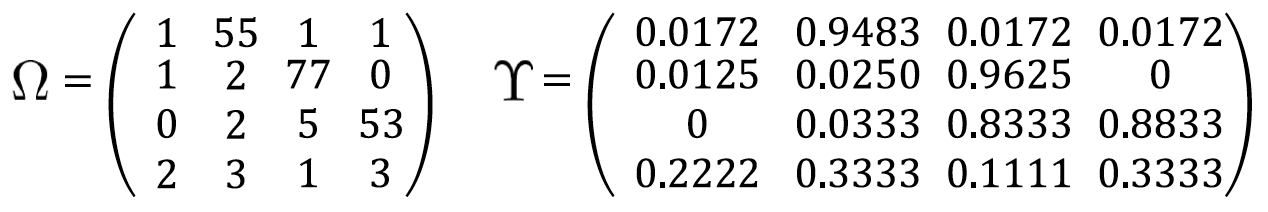}
\caption{
Visualization of auxiliary matrix $\Omega$ and transition matrix $\Upsilon$ of the task ``Replace a Bulb'', which contains 4 steps as ``remove the light shell'', ``take out the old bulb'', ``install the new bulb'' and ``install the light shell''.}
\label{fig:vis_mat}
\end{figure}

Here $\textbf{f}^n(t)$ denotes the probabilities of $K$ steps occurring at the time-stamp $t$, and we set $\mu_n = (t^{n1}_{in} + t^{n2}_{in})/2, \sigma_n = (t^{n2}_{in} - t^{n1}_{in})/2$ respectively.
For a whole video, we obtain the function of step score as $\textbf{f}(t) = \sum_n \textbf{f}^n(t)$.
Then, at time-stamp t, we sum the scores of all the steps to calculate its corresponding binary actionness score $a(t)$.
After that, we follow~\cite{DBLP:conf/iccv/ZhaoXWWTL17} to apply the watershed algorithm on the 1D signal $a(t)$ to obtain a sequence of segments $S = \{\textbf{t}^l_s,\textbf{s}^l_s\}_{l=1}^L$,
where its validness is theoretically guaranteed by~\cite{DBLP:journals/fuin/RoerdinkM00}.
Here, $\textbf{t}^l_s = [t^{l1}_s,t^{l2}_s]$ is the temporal interval of the segment $S^l$,
where we calculate the step score as:
\begin{eqnarray}
\textbf{s}_s^l = \int_{t_s^{l1}}^{t_s^{l2}} \textbf{f}(t)\mathrm{d}t.
\end{eqnarray}

\textbf{Refining Segments S by Ordering Constraint.}
In order to refine the S by the intrinsic ordering-dependency in instructional video, 
we leverage a transition matrix $\Upsilon$ where the element $\Upsilon_{ij}$ of the matrix denotes the probability of step $j$ follows step $i$: 
\begin{eqnarray}
   \Upsilon_{ij} = p(S^l = j | S^{l-1} = i), \quad l = 2,3,..,L.
\end{eqnarray}

To construct $\Upsilon$,
we first introduce an auxiliary matrix $\Omega$ by counting the occurrence time of the step $j$ after step $i$ based on the all ordered step lists appearing in the training set:
\begin{eqnarray}
    \Omega_{ij} = \#\textit{(step j follows step i).}
\end{eqnarray}

We normalize each row of $\Omega$ to obtain the transition matrix $\Upsilon$.
The $\Omega$ and $\Upsilon$ of a task is presented in Fig. \ref{fig:vis_mat}, and more examples can be found in the supplementary.
Then, for each list of segments $\{ s_s^l \}_{l=1}^L$,
we refine their scores as:
\begin{eqnarray}
   \textbf{s}_{s,new}^l = 
    \begin{cases}
\lambda_1 \textbf{s}_s^1 + \lambda_2 \bm{\eta} , \quad l = 1.\\
\lambda_1 \textbf{s}_s^l + \lambda_2 \Upsilon \textbf{s}_{s,new}^{l-1}, \quad l = 2, 3, ..., L. \\
    \end{cases}
\end{eqnarray}
where $\lambda_1$ and $\lambda_2$ ($\lambda_1 + \lambda_2 = 1, \lambda_1, \lambda_2 > 0 $) are two hyper-parameters to balance the effects of the original score
and the ordering regularized score.
$\bm{\eta}$ is the probability of different steps occurring at the first segment:
\begin{eqnarray}
  \bm{\eta}_i = \frac{\#\textit{(step i occurs at the first segment)}}{\#\textit{(total videos)}}.
\end{eqnarray}

In Eqn. (9),
we borrow idea from
hidden Markove Model (HMM)~\cite{baum1966statistical}, 
which mathematically defined the state transition equation as
$\boldsymbol{\pi}^t = \boldsymbol{P} \boldsymbol{\pi}^{t-1}$. Here $\boldsymbol{\pi}^t,\boldsymbol{\pi}^{t-1}$ and $\boldsymbol{P}$ correspond to the segment scores and transition matrix in our paper.
Besides, in order to combine the orignal score and the ordering regularized score, we utilize the score fusion method,
which is a simple yet effective scheme and similarly adopted in previous works~\cite{two-stream,TSN2016ECCV} empirically.
We also evaluate other fusion strategies in section 5.1 experimentally.

\textbf{Mapping Variation $\Delta \textbf{s}_s^l$ to $\Delta \textbf{s}_{in}^n$ .}
The ``mapping variation'' can be regarded as an inverse process of ``grouping proposals'', 
which maps the variation in the segment space $\mathbb{S}$ into the original proposal space $\mathbb{P}$ for the later evaluation process.
Specifically, for a region $\textbf{t}_s^l$, we obtain the variation of $\Delta \textbf{s}_s^l$ as $\Delta \textbf{s}_s^l = \textbf{s}_{s,new}^l - \textbf{s}_s^l$.
Then we have the $\Delta \textbf{f}(t)$ as:
\begin{eqnarray}
    \Delta \textbf{f}(t) = 
    \begin{cases}
\Delta \textbf{s}_s^l ./ \textbf{s}_s^l \odot \textbf{f}(t), \quad \textit{if}  \;  \; \exists  \; \textbf{t}_s^l, s.t. \; t \in \textbf{t}_s^l.\\
0, \quad \textit{otherwise}. \\
    \end{cases}
\end{eqnarray}

Here $./$ and $\odot$ denote element-wise division and production, and we assume the ratio of the variation $\Delta \textbf{s}_s^l$ and $\Delta \textbf{f}(t)$ equal to the ratio between their original value.
Similarly, we calculate the variation $\Delta \textbf{s}_n$ and refine the proposal score $\textbf{s}_n$ as follow:
\begin{eqnarray}
\Delta \textbf{s}_{in}^n &=& \int_{0}^T \Delta \textbf{f}^n(t) \mathrm{d}t = \int_{0}^T  \Delta \textbf{f}(t) ./ \textbf{f}(t) \odot \textbf{f}^n(t) \mathrm{d}t. \\
\textbf{s}_{out}^n &=& \textbf{s}_{in}^n + \Delta \textbf{s}_{in}^n, \quad \textbf{t}_{out}^n = \textbf{t}_{in}^n.
\end{eqnarray}

In practice,
we multiply a Gaussian function $\varphi(t;\mu_n,\sigma_n)$ as a regularized factor before the integration.
This operation concentrates more energy to the middle of the proposal and is shown to be more effective during the experiments.
As there are several time-wise functions like $\textbf{f}(t)$ and $\textbf{f}^n(t)$, we discretize the time into $M$ slots and use accumulation to approximate the continuous integration. We set $M$ to 100 in our experiments empirically.
Having obtained the output proposals $P_{out}=\{\textbf{t}_{out}^n,\textbf{s}_{out}^n\}_{n=1}^N$ which are regularized by the prior ordering-dependency knowledge, 
we perform NMS to obtain the final step localization results.

\linespread{1.1}
\begin{table*}[ht]
\small
\caption{Comparisons of the step localization accuracy (\%) of the baselines and our task-consistency (TC) method on the COIN dataset.} \label{tab:coin_tc}
\setlength{\tabcolsep}{9.5pt}
\centering
\begin{tabular}{l | c c c c c | c c c c c }
\toprule[1.5pt]
 & \multicolumn{5}{|c|}{ mAP @ $\alpha$} &
\multicolumn{5}{|c}{ mAR @ $\alpha$}\\
Method & 0.1 & 0.2 & 0.3 & 0.4 & 0.5
& 0.1 & 0.2 & 0.3 & 0.4 & 0.5\\
\midrule[1.2pt] 
Random & 0.03 & 0.03 & 0.02 & 0.01 & 0.01 & 2.57 & 1.79 & 1.36& 0.90 & 0.50\\
\hline
SSN$_{\textit{-RGB}}$\cite{DBLP:conf/iccv/ZhaoXWWTL17} & 19.39 & 15.61 & 12.68 & 9.97 & 7.79 & 50.33 & 43.42 & 37.12 & 31.53 & 26.29\\
SSN$_{\textit{-Flow}}$\cite{DBLP:conf/iccv/ZhaoXWWTL17} &  11.23 & 9.57 & 7.84 & 6.31 & 4.94 & 33.78 & 29.47 & 25.62 & 21.98 & 18.20\\
SSN$_{\textit{-Fusion}}$\cite{DBLP:conf/iccv/ZhaoXWWTL17} &  20.00 & 16.09 & 13.12 & 10.35 & 8.12 & 51.04 & 43.91 & 37.74 & 32.06 & 26.79\\
\hline
SSN+TC$_{\textit{-RGB}}$ & 20.15 & 16.79 & 14.24 & 11.74 & 9.33 & 54.05 & 47.31 & 40.99 & 35.11 & 29.17 \\
SSN+TC$_{\textit{-Flow}}$ &  12.11 & 10.29 & 8.63 & 7.03 & 5.52 & 37.24 & 32.52 & 28.50 & 24.46 & 20.58 \\
SSN+TC$_{\textit{-Fusion}}$ &  20.01 & 16.44 & 13.83 & 11.29 & 9.05 & 54.64 & 47.69 & 41.46 & 35.59 & 29.79\\
\bottomrule[1.5pt]
\end{tabular}
\end{table*}
\linespread{1}

\section{Experiments}
In this section,
we study five different tasks for instructional video analysis: (1) step localization, (2) action segmentation, (3) proposal localization, (4) task recognition and (5) step recognition\footnote{We present a table to clarify the goal, metric, and evaluated methods for each task in supplementary material.}.
For the first two important tasks in this field, we evaluate various existing approaches to provide a benchmark on our COIN dataset.
For the other three tasks, our goal is to compare the challenge of our COIN with other relative datasets based on the same method.
We also test our task-consistency and ordering-dependency methods for the step localization tasks.
In addition, we further conduct experiments to evaluate whether COIN can help for training models on other instructional datasets.
The following describes the details of our experiments and results.

\subsection{Evaluation on Step Localization}
\textbf{Implementation Details:}
In this task, we aim to localize a series of steps and recognize their corresponding labels given an instructional video.
We mainly evaluate the following approaches:
(1) Random. We uniformly segmented the video into three intervals, and randomly assigned the label to each interval.  
(2) SSN\cite{DBLP:conf/iccv/ZhaoXWWTL17}.
This is an effective model for action detection, 
which outputs the same type of results (interval and label for each action instance) with step localization.
We utilized the PyTorch implementation
and the BNIcenpention as the backbone. 
We followed the default setting to sample 9 snippets from each segment.
We used the SGD optimizer to train the model with the initial learning rate of 0.001.
The training process lasted for 24 epochs, and the learning rate was scaled down by 0.1 at the 10th and 20th epoch.
The NMS threshold was set to 0.6.
The reported results are based on the inputs of different modalities as:
SSN$_{-RGB}$, SSN$_{-Flow}$ and SSN$_{-Fusion}$. 
Here SSN$_{-Flow}$ adopted the optical flows calculated by ~\cite{zach2007duality},
and SSN$_{-Fusion}$ combined the predicted scores of SSN$_{-RGB}$ and SSN$_{-Flow}$.
(3) SSN+TC, SSN+OD, SSN+ODTC, SSN+TCOD. 
We test these methods in order to demonstrate the advantages of the proposed method to explore the task-consistency (TC) and ordering-dependency (OD) in instructional videos.
For clarification, +ODTC denotes first performing ordering-dependency regularization then executing task-consistency method, while +TCOD is the other way round. 
(4) R-C3D\cite{DBLP:conf/iccv/XuDS17}, BSN\cite{BSN} and BMN~\cite{BMN} with TC and OD.
We further plugged our TC and OD methods into these action detection models to verify their generalization ability.
Since the BMN and BSN were originally designed for temporal action proposal generation, 
we processed the proposals generated by them with the classifier of SSN to produce the final results.

\begin{figure*}[tb]
\setlength{\abovecaptionskip}{-0.2cm}
\includegraphics[width = \linewidth]{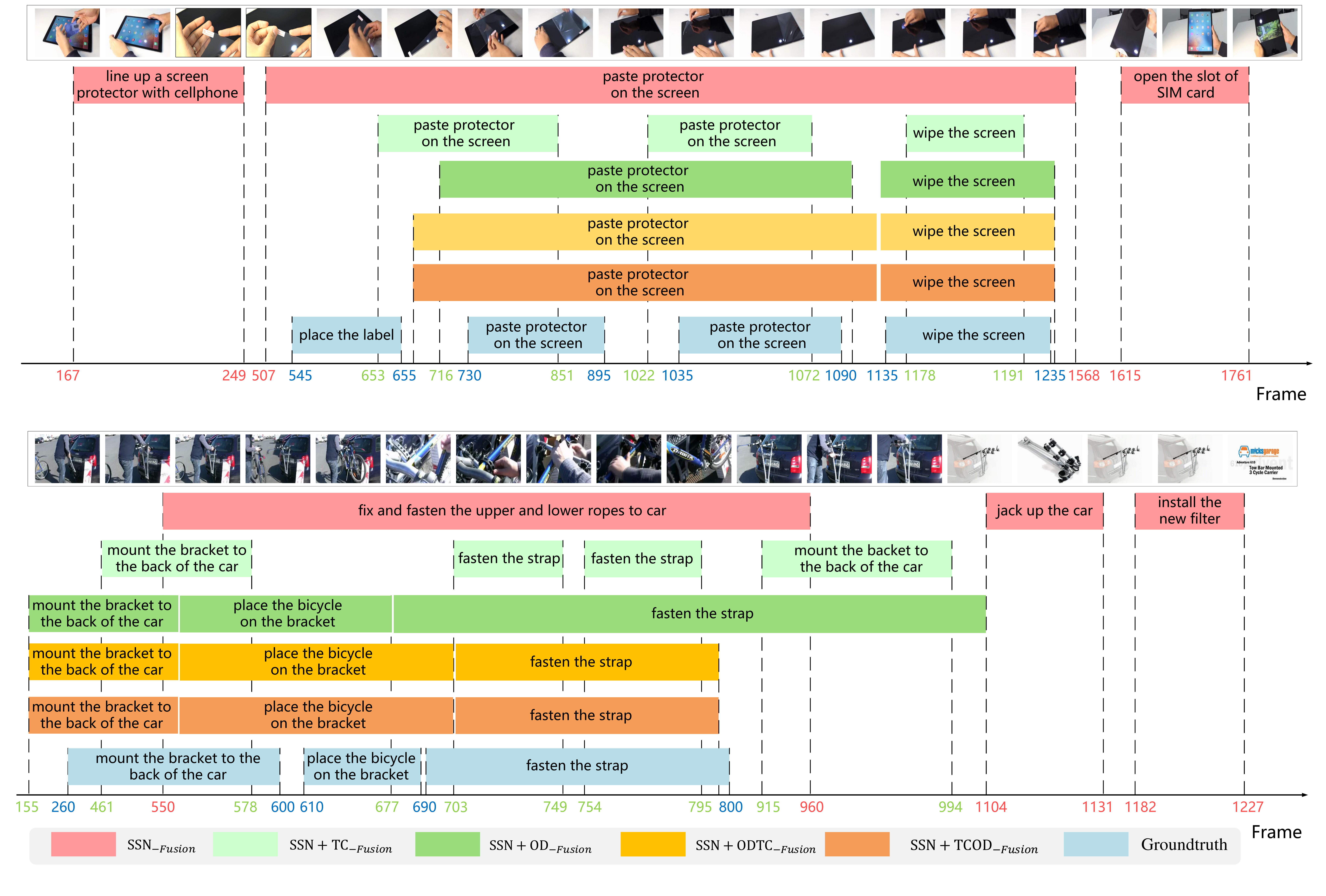}
\caption{Visualized results of step localization. 
The videos belongs to the tasks of ``paste screen protector on Pad'' and ``install the bicycle rack''.
}
\label{fig:vis_sl}
\vspace{-0.1cm}
\end{figure*}

\textbf{Evaluation Metrics:}
As the results of step localization contain time intervals, labels and confidence scores, 
we employed Intersection over Union (IoU) as a basic metric to determine whether a detected interval is positive or not.
The IoU is defined as $|G\cap D| / |G\cup D|$, where G denotes the ground truth action interval and D denotes the detected action interval. 
We followed \cite{DBLP:journals/corr/LiuHLSL17} to calculate \textit{Mean Average Precision (mAP)} and \textit{Mean Average Recall (mAR)}.
The results are reported under the IoU threshold $\alpha$ ranging from 0.1 to 0.5.

\textbf{Baseline Results:}
Table \ref{tab:coin_tc} presents the experimental results
which reveal great challenges to performing step localization on the COIN dataset.
Even for the state-of-the-art method SSN$_{-Fusion}$, 
it only attains the results of 8.12\% and 26.79\% on mAP@0.5 and mAR@0.5 respectively.

\textbf{Analysis on Task-consistency Method:}
For the task-consistency method, we observe that SSN+TC improves the performance over the baseline models in Table \ref{tab:coin_tc},
which illustrates the effectiveness of our proposed method to capture the dependencies among different steps.
We show several visualization results of different methods and ground-truth in Fig. \ref{fig:vis_sl}.
As an example, we analyze an instructional video of the task ``paste screen protector on Pad'' as follow.
When applying our task-consistency method, 
we can discard those steps which do not belong to this task, \textit{e.g.,} ``line up a screen protector with cellphone'' and ``open the slot of SIM card'',
hence more accurate step labels can be obtained.

As we discussed in Section 4.1,
during the bottom-up aggregation period,
there are two approaches to calculate the task score: (1) summing the scores of steps belonging to the task, or (2) averaging them according to the number of steps.
We display the compared results in Table \ref{tab:sum_ave}, 
the summing strategy achieves better performance than the averaging method under both mAP and mAR metrics.
Actually, they equally weigh all the steps or tasks respectively,
and the summing method is shown to be more effective on the COIN.

\linespread{1.2}
\begin{table}[t]
\centering 
\caption{
Study of different approaches for calculating the task score in task-consistency method on the COIN dataset.
}
\begin{tabular}{c||c|c|c}
\toprule[1.5pt]
 & SSN$_{-RGB}$ & +TC(sum) & +TC(ave) \\
\hline
 mAP@0.5 & 7.79 & 9.33 & 9.16 \\
 mAR@0.5 & 26.29 & 29.17 & 29.02 \\
\bottomrule[1.5pt]
\end{tabular}
\label{tab:sum_ave}
\vspace{-0.2cm}
\end{table}
\linespread{1}

\linespread{1.1}
\begin{table}[t]
\small
\caption{\footnotesize Analysis on our proposed TC and OD methods on the COIN dataset. The results are based on mAP under the IoU threshold $\alpha = 0.1$.
} 
\centering
\begin{tabular}{l || c c c }
\toprule[1.5pt]
 & RGB &  Flow & Fusion\\
\hline
SSN\cite{DBLP:conf/iccv/ZhaoXWWTL17} & 19.39 & 11.23 & 20..00\\
SSN+TC & 20.15 & 12.11&20.01\\
SSN+OD & 20.39 & 11.82&20.95\\
SSN+TCOD & 21.82 & 13.09&21.77\\
SSN+ODTC & 23.23 & 13.25&22.89\\
\bottomrule[1.5pt]
\end{tabular}
\label{tab:coin_tcod}
\vspace{-0.2cm}
\end{table}
\linespread{1}

\textbf{Analysis on Ordering-dependency Method:}
As shown in Table \ref{tab:coin_tcod}, our proposed ordering-dependency (OD) method surpasses the baseline model on both mAP and mAR metrics,
which verifies its effectiveness for the step localization task.
Moreover, when combining with the task-consistency (TC) method, it can achieve further improvements.
Specifically, the ODTC (first performing OD then executing TC) achieves the best result of 23.23\% accuracy on RGB modality, slightly outperforming the TCOD (first performing TC then executing OD).
Some visualization results have also been shown in Fig. \ref{fig:vis_sl}.
In most cases, our proposed methods can obtain more accurate results.
However in the first task, for the SSN+ODTC$_{-Fusion}$ and SSN+TCOD$_{-Fusion}$, 
the results are little worse than the result of SSN+OD$_{-Fusion}$ due to the nebulous boundary of some steps.

\linespread{1.2}
\begin{table}[t]
\centering 
\caption{
Study of the hyper-parameters $\lambda_1$ and $\lambda_2$ for step localization on the COIN and Breakfast dataset.
The results are all based on SSN+OD$_{\textit{-Fusion}}$ under mAP@0.1.
}
\begin{tabular}{c c c c c c c }
\toprule[1.5pt]
$\lambda_1$ & 0 & 0.2& 0.4 & 0.6& 0.8  & 1 \\
$\lambda_2$ & 1 & 0.8 & 0.6&0.4 & 0.2& 0 \\
\hline
COIN & \textbf{20.95} & 20.76 & 20.58 &20.44 & 20.27 & 20.01 \\
Breakfast & 26.71 & 27.61 & 28.17 &28.71 & \textbf{29.05} &  28.24 \\
\bottomrule[1.5pt]
\end{tabular}
\label{tab:lambda}
\vspace{-0.2cm}
\end{table}

Besides, in ordering-dependency method, there are two hyper-parameters $\lambda_1$ and $\lambda_2$, 
which make a trade-off for the original score and ordering regularized score.
We explore different $\lambda_1$ and $\lambda_2$ on both COIN and Breakfast datasets and present the results in Table \ref{tab:lambda}.
We observe that in COIN dataset, the peak reaches at $\lambda_1 = 0, \lambda_2 =1$, suggesting that the ordering-dependency information is more important.
However for the Breakfast dataset, we achieve the best result at $\lambda_1 = 0.8, \lambda_2 =0.2$, which indicates that the original score obtained based on the appearance information contributes more.

In Eqn (5), we use a Gaussian based function for an input proposal.
Here we study other distributions with different standard deviation $\sigma_n = \beta \cdot  \frac{t^{n2}_{in} - t^{n1}_{in}}{2}, 
$
where $\beta$ is a standard deviation factor.
Besides, we explore another Triangle distribution: 
\begin{eqnarray}
   \textbf{f}^n(t) = 
    \begin{cases}
\frac{|t-\mu_n|}{t^{n2}_{in} - t^{n1}_{in}}\textbf{s}^n_{in}, \quad t^{n1}_{in} \leq t \leq t^{n2}_{in},\\
\textbf{0}, \quad else. \\
    \end{cases}
\end{eqnarray}

We show the experimental results in Table \ref{tab:od_ab}, which indicates that the Gaussian based distribution with the standard deviation factor $\beta = 1$ is a proper choice.

In Eqn (9), we adopt the weight score fusion method to refine the segment score, while here we explore other different methods. 
Specifically, we denote the original score and the
ordering regularized score as follow:
\begin{eqnarray}
\textbf{s}_1 = \textbf{s}_s^l,  
\quad \textbf{s}_2 = 
    \begin{cases}
\bm{\eta}, \quad l = 1,\\
\Upsilon \textbf{s}_{s,new}^{l-1}, \quad l = 2, 3, ..., L. \\
    \end{cases}
\end{eqnarray}

\linespread{1.2}
\begin{table}[t]\footnotesize
\caption{Analysis of the OD method.
Experiments are conducted on the RGB modality of the COIN dataset. G and T denote the Gaussian based distribution and Triangle distribution respectively, while
the second element in parentheses denotes the standard deviation factor $\beta$.
See text for the definitions of other variables and more details.
}
\centering
\begin{tabular}{c | c c c c c }
\toprule[1.5pt]
\multicolumn{6}{c}{Study of different generated distributions for calculating scores}  
 \\
\toprule[1pt]
Distribution & (G,0.5) & (G,1) & (G,2)
& (G,5) & T  \\
mAP@0.1 & 20.07 &  20.39 & 20.13
& 19.81 & 19.77  \\
\end{tabular}
\begin{tabular}{c | c c c c }
\toprule[1pt]
 \multicolumn{5}{c}{Study of different methods to refine the segment scores ($\lambda_1 = \lambda_2 = 0.5$)} \\
\toprule[1pt]
Method & $\lambda_1 \textbf{s}_1 + \lambda_2 \textbf{s}_2$ & $\sqrt{\lambda_1 \textbf{s}_1^2 + \lambda_2 \textbf{s}_2^2}$ & $\textbf{s}_1^{\lambda_1} \textbf{s}_2^{\lambda_2}$  & max-pool$(\textbf{s}_1, \textbf{s}_2)$ \\
mAP@0.1 & 20.30 &  20.09 & 20.36
& 19.84   \\
\toprule[1pt]
 \multicolumn{5}{c}{Study of the number of time slots M} \\
\toprule[1pt]
M & 50 &  100 & 150
& 200\\
mAP@0.1 & 19.87 & 20.39 & 20.55 & 20.58\\
\bottomrule[1.5pt]
\end{tabular}
\vspace{-0.1cm}
\label{tab:od_ab}  
\end{table}

We explore different approaches to fuse these two scores. Table \ref{tab:od_ab} shows that ``$\lambda_1 \textbf{s}_1 + \lambda_2 \textbf{s}_2$''
and ``$\textbf{s}_1^{\lambda_1} \textbf{s}_2^{\lambda_2}$'' are two methods which can achieve better results for refining the segment scores.

Moreover, we futher present the evalutation results of the time slot M in Table \ref{tab:od_ab}.
It can be seen that finer-grained division with larger M can lead to better performance.
In this paper, we use $M=100$ to make a good trade-off between the effectiveness and efficiency as larger M would also bring more computational cost.

\begin{table}[t]
\small
\caption{\footnotesize Comparisons of the step localization accuracy (\%) over 12 domains on the COIN dataset. 
We report the results obtained by SSN+TC$_{-Fusion}$ with $\alpha$ = 0.1.}
\centering
\begin{tabular}{ c c | c  c  }
\toprule[1.5pt]
Domain & mAP & Domain & mAP \\
\hline
nursing \& caring  &  22.92  & vehicles & 19.07 \\
science \& craft   &  16.59  & electric appliances  & 19.86  \\
leisure \& performance  &  24.32  & gadgets & 17.99  \\
snacks \& drinks  &  19.79  & dishes & 23.76  \\
plants \& fruits  &  22.71  & sports & 30.20  \\
household items  &  19.07  & housework & 20.70  \\
\bottomrule[1.5pt]
\end{tabular}
\vspace{-0.2cm}
\label{tab:ex_domain}
\end{table}
\linespread{1}

\textbf{Discussion:} We provide some further discussions as below.

(1) \textit{What are the hardest and easiest domains for instructional video analysis?}
In order to provide a more in-depth analysis of the COIN dataset, 
we report the performance of SSN+TC$_{-Fusion}$ among the 12 domains.
Table \ref{tab:ex_domain} presents the comparison results,
where the domain ``sports'' achieves the highest mAP of 30.20\%, 
This is because the differences between the ``sports'' steps are more clear,
thus they are easier to be identified.
In contrast, the results of ``gadgets'' and ``science \& craft'' are relatively low.
The reason is that the steps in these two domains usually have higher similarity with each other.
For example, the step ``remove the tape of the old battery'' is similar to the step ``take down the old battery''.
Hence it is harder to localize the steps in these two domains.
For the compared performance across different tasks, please see in the supplementary material for more details.

(2) \textit{Can the proposed task-consistency and ordering-dependency methods be applied to other action detection models?}
Since our proposed TC and OD are two plug-and-play methods, we futher validate them on the R-C3D~\cite{DBLP:conf/iccv/XuDS17}, BSN~\cite{BSN} and BMN~\cite{BMN} models.
From Table \ref{tab:othermodel_tcod} we can see that both TC and OD could improve the performance of various basic models, which further demonstrate the effectiveness of our proposed methods.

\linespread{1.1}
\begin{table}[t]\small
\centering 
\caption{
Study of the TC and OD approaches on different basic models. The results are reported based on the RGB modality in the COIN dataset under mAP@0.1 (\%).
}
\begin{tabular}{c c c c}
\toprule[1.5pt]
Basic Model & R-C3D~\cite{DBLP:conf/iccv/XuDS17} & BSN~\cite{BSN} & BMN~\cite{BMN} \\
\hline
Baseline & 9.85 & 18.91 &18.60  \\
Baseline+TC  & 10.32 & 19.96& 19.27 \\
Baseline+OD & 10.08& 20.46& 19.68\\
\bottomrule[1.5pt]
\end{tabular}
\vspace{-0.2cm}
\label{tab:othermodel_tcod}
\end{table}

\linespread{1}

(3) \textit{Can the proposed task-consistency and ordering-dependency methods be applied to other instructional video datasets?}
In order to demonstrate the effectiveness of our proposed methods,
we further conduct experiments on another dataset called ``Breakfast"\cite{DBLP:conf/cvpr/KuehneAS14},  which is also widely-used for instructional video analysis. 
The Breakfast dataset contains over 1.9k videos with 77 hours of 4 million frames. Each video is labelled with a subset of 48 cooking-related action categories. 
Following the default setting, we set split 1 as testing set and the other splits as training set.
Similar to COIN,
we employ SSN\cite{DBLP:conf/iccv/ZhaoXWWTL17}, which is a state-of-the-art method for action detection, 
as a baseline method under the setting of step localization.
As shown in Table \ref{tab:break_tcod}, 
our proposed task-consistency and ordering-dependency methods improve the performance of the baseline model,
which further shows their advantages to model the dependencies of different steps in instructional videos.
Besides, the ODTC achieves better results than the TCOD, which illustrates that first performing OD is more effective.

\linespread{1.2}
\begin{table}[t]
\small
\caption{\footnotesize Comparisons of the step localization accuracy (\%) on the Breakfast dataset. 
The results are all based on the combination scores of RGB frames and optical flows.
} 
\centering
\begin{tabular}{l | c c c | c c c }
\toprule[1.5pt]
Metrics & \multicolumn{3}{c}{mAP@ $\alpha$} &
\multicolumn{3}{|c}{mAR@ $\alpha$}\\
Threshold & 0.1 &  0.3 & 0.5
& 0.1& 0.3  & 0.5\\
\hline
SSN\cite{DBLP:conf/iccv/ZhaoXWWTL17} &  28.24 & 22.55 & 15.84 & 54.86 & 45.84 & 35.51\\
SSN+TC  &  28.25 & 22.73 & 16.39 & 55.51 & 47.37 & 36.20 \\
SSN+OD  &  29.05 & 22.61 & 15.90 & 55.81 & 45.85 & 35.79 \\
SSN+TCOD  & 30.89 &24.35 &16.87 &57.80&48.84 &36.51 \\
SSN+ODTC  & 30.91 &24.45 &16.94 &57.91&49.33 &36.78 \\
\bottomrule[1.5pt]
\end{tabular}
\vspace{-0.2cm}
\label{tab:break_tcod}
\end{table}
\linespread{1}

\subsection{Evaluation on Action Segmentation}
\textbf{Implementation Details:}
The goal of this task is to assign each video frame with a step label. 
We present the results on three types of approaches as follows.
(1) Random. We randomly assigned a step label to each frame. 
(2) Fully-supervised method. We used VGG16 network pre-trained on ImageNet,
and finetuned it on the training set of COIN to predict the frame-level label. 
(3) Weakly-supervised approaches. 
In this setting, we evaluated recent proposed Action-Sets\cite{richard2017action}, NN-Viterbi\cite{DBLP:journals/corr/abs-1805-06875} and TCFPN-ISBA\cite{DBLP:journals/corr/abs-1803-10699} without temporal supervision. 
For Action-Sets, only a set of steps within a video is given, while the occurring order of steps are also provided for NN-Viterbi and TCFPN-ISBA. 
We used frames or their representations sampled at 10fps as input.
We followed the default train and inference pipeline of Action-Sets\cite{richard2017action}, NN-Viterbi\cite{DBLP:journals/corr/abs-1805-06875} and \cite{DBLP:journals/corr/abs-1803-10699}.
However, these methods use frame-wise fisher vector as video representation,
which comes with huge computation and storage cost on the COIN dataset\footnote{ The calculation of fisher vector is based on the improved Dense Trajectory (iDT) representation~\cite{wang2013action}, which requires huge computation cost and storage space.}. 
To address this,
we employed a bidirectional LSTM on the top of a VGG16 network to extract dynamic feature of a video sequence\cite{DBLP:journals/pami/DonahueHRVGSD17}.

\textbf{Evaluation Metrics:}
We adopted frame-wise accuracy (FA), which is a common benchmarking metric for action segmentation.
It is computed by first counting the number of correctly predicted frames, 
and dividing it by the number of total video frames.

\linespread{1.2}
\begin{table}[tb]\small
\setlength{\abovecaptionskip}{-0.02cm}
\caption{\footnotesize Comparisons of the action segmentation accuracy (\%) on the COIN.}
\centering
\begin{tabular}{l | c | c}
\toprule[1.5pt]
Method & Frame Acc. & Setting\\
\hline
Random & 0.13 & - \\ 
\hline
CNN~\cite{Karen15very} & 25.79& fully-supervised\\
\hline
Action-Sets\cite{richard2017action} &  4.94 & weakly-supervised\\
NN-Viterbi\cite{DBLP:journals/corr/abs-1805-06875} & 21.17 & weakly-supervised\\
TCFPN-ISBA\cite{DBLP:journals/corr/abs-1803-10699} & 34.30 & weakly-supervised\\
\bottomrule[1.5pt]
\end{tabular}
\label{tab:action_seg}
\end{table}
\linespread{1}

\textbf{Results: }Table \ref{tab:action_seg} shows the results of action segmentation on the COIN. 
Given the weakest supervision of video transcripts without ordering constraint,
Action-Sets\cite{richard2017action} achieves the result of 4.94\% frame accuracy. 
When taking into account the ordering information, 
NN-Viterbi~\cite{DBLP:journals/corr/abs-1805-06875} and TCFPN-ISBA~\cite{DBLP:journals/corr/abs-1803-10699}
outperform Action-Sets with a large margin of 16.23\% and 29.66\% respectively. 
As a fully-supervised method, CNN~\cite{Karen15very} reaches an accuracy 25.79\%, 
which is much higher than Action-Sets.
This is because CNN utilizes the label of each frame to perform classification and the supervision is much stronger than Action-Sets.
However, as the temporal information and ordering constraints are ignored,
the result of CNN is inferior to TCFPN-ISBA.

\subsection{Comparison with Other Video Analysis Datasets}
\linespread{1.2}
\begin{table}[tb]
\setlength{\abovecaptionskip}{-0.02cm}
\centering 
\caption{Comparisons of the proposal localization accuracy (\%) with YouCook2 dataset~\cite{DBLP:conf/aaai/ZhouXC18}.
The results are obtained by temporal actionness grouping (TAG) method~\cite{DBLP:conf/iccv/ZhaoXWWTL17} with $\alpha$ = 0.5.}
\small
\begin{tabular}{ c c c|c c c}
\toprule[1.5pt]
& YouCook2 & COIN & & YouCook2 & COIN \\
\hline
mAP & 38.05 & 39.67 & mAR & 50.04 & 56.16\\
\bottomrule[1.5pt]
\end{tabular}
\label{tab:sl_wyoucook2}
\vspace{-0.2cm}
\end{table}
\linespread{1}
In order to assess the difficulty of COIN, 
we report the performance on different tasks compared with other datasets. 

\textbf{Proposal Localization:}
As defined in~\cite{DBLP:conf/aaai/ZhouXC18},
proposal localization aims to segment an instructional video into category-independent procedure segments.
For this task,
we evaluated COIN and YouCook2~\cite{DBLP:conf/aaai/ZhouXC18} based on temporal actionness grouping (TAG) approach~\cite{DBLP:conf/iccv/ZhaoXWWTL17}.
We follow the experiments setting in~\cite{DBLP:conf/aaai/ZhouXC18} to generate 10 proposals for each videos,
and report the results of mAR@0.5 and mAP@0.5.
From the results in Table \ref{tab:sl_wyoucook2}, 
we observe that the mAP and mAR of the same method are lower on the YouCook2 dataset,
which indicates that it is more challenging than the COIN for the proposal localization task.

\textbf{Video Classification:}
For video classification on COIN,
we conducted experiments on task recognition and step recognition.
The task recognition takes a whole video as input, and predicts the \textit{task} label referring to the second level of the lexicon.
The step recognition takes the trimmed segments as input, and outputs the \textit{step} label corresponding to the third level of the lexicon.
We employed the temporal segment network (TSN) model~\cite{TSN2016ECCV},
which is a state-of-the-art method for video classification.
As shown in the Table \ref{tab:coin_vc},
the task recognition accuracy on COIN is 73.36\%,
suggesting its general difficulty in comparison with other datasets.
However, the step recognition accuracy is only 36.46\%, as it requires discriminating different steps at finer level.
Besides, since the step recognition task is performed on the clips trimmed by the ground-truth temporal intervals,
its accuracy can be considered as a reference for step localization which is a highly related and more complex task.

\textbf{Step Localization:}
For action detection or step localization,
we display the compared performances of structured segment networks (SSN) approach~\cite{DBLP:conf/iccv/ZhaoXWWTL17} on COIN and the other three datasets in Table \ref{tab:cmp_data_det}.
The THUMOS14~\cite{THUMOS14} and ActivityNet~\cite{DBLP:conf/cvpr/HeilbronEGN15} are conventional datasets for action detection,
on which the detection accuracies are relatively higher.
The Breakfast~\cite{DBLP:conf/cvpr/KuehneAS14} and COIN contain instructional videos with more difficulty.
Hence, the performance on these two datasets are lower.
Especially for our COIN, the results of mAP@0.5 is only 8.12\%. We attribute the low performance to two aspects:
(1) The step intervals are usually shorter than action instances, which brings more challenges for temporal localization; 
(2) Some steps in the same tasks are similar,
which carry ambiguous information for the recognition process.
These two phenomena are also common in real-world scenarios,
and future works are encouraged to address these two issues.

\linespread{1.2}
\begin{table}[tb]
\setlength{\abovecaptionskip}{-0.02cm}
\caption{Comparisons of the video classification performance (\%) on different datasets.
The reported results are based on temporal segment networks (TSN) model~\cite{TSN2016ECCV}.} 
\centering
\begin{tabular}{l c }
\toprule[1.5pt]
Dataset & Accuracy \\
\hline
UCF101~\cite{UCF101} & 97.00 \\
ActivityNet v1.3~\cite{DBLP:conf/cvpr/HeilbronEGN15} & 88.30   \\
Kinectics~\cite{DBLP:conf/cvpr/CarreiraZ17} & 73.90 \\
\hline
COIN (task recognition) & 73.36 \\
COIN (step recognition) & 36.46 \\
\bottomrule[1.5pt]
\label{tab:coin_vc}
\end{tabular}
\vspace{-0.2cm}
\end{table}
\linespread{1}

\linespread{1.2}
\begin{table}[tb]
\setlength{\abovecaptionskip}{-0.02cm}
\caption{Comparisons of the action detection/step localization performance (\%) on different datasets.
The reported results are based on stuctured segment networks (SSN) method~\cite{DBLP:conf/iccv/ZhaoXWWTL17} with $\alpha$ = 0.5.} \label{tab:cmp_data_det}
\centering
\begin{tabular}{l c }
\toprule[1.5pt]
Dataset & mAP   \\
\hline
 THUMOS14~\cite{THUMOS14} &  29.10 \\
 ActivityNet v1.3~\cite{DBLP:conf/cvpr/HeilbronEGN15} & 28.30\\
  Breakfast~\cite{DBLP:conf/cvpr/KuehneAS14} & 15.84\\
  \hline
  COIN & 8.12\\
\bottomrule[1.5pt]
\end{tabular}
\vspace{-0.2cm}
\end{table}
\linespread{1}

\subsection{Cross Dataset Transfer}
In order to see whether the COIN could benefit other instructional video datasets from different domains,
we further study the cross dataset transfer setting.
Under the ``pre-training + fine-tuning'' paradigm, we conducted experiments to verify the step localization task on the Breakfast\cite{DBLP:conf/cvpr/KuehneAS14}, JIGSAWS\cite{gao2014jhu} and UNLV-Diving\cite{DBLP:conf/cvpr/ParmarM17} datasets.
The Breakfast dataset is based on cooking activity while the JIGSAWS dataset consists of three surgical tasks.
In comparison, the UNLV-diving dataset~\cite{DBLP:conf/cvpr/ParmarM17} contains a completely different task (diving) in a very different environment (natatorium) compared with the tasks in COIN.
This dataset is originally collected for action quality assessment, 
where we employ the annotation of three steps as ``jumping'', ``dropping'' and ``entering into water'' to perform step localization.

For the Breakfast and the UNLV-Diving dataset, we split the training and testing set following\cite{DBLP:conf/cvpr/KuehneAS14,DBLP:conf/cvpr/ParmarM17}.
For the JIGSAWS dataset, we evaluate the performance on the split 1 of the leave-one-user-out setting suggested in~\cite{gao2014jhu}.
In the cross dataset transfer experiments, we used two pre-trained models on ``Kinetics'' and ``Kinetics + COIN'',
and fine-tuned the model on the target datasets.
Since the Breakfast and JIGSAWS datasets are both self-collected and the UNLV-Diving dataset is created from the professional sport videos,
there will not be any duplicate videos from the COIN and Kinetics which are sourced from YouTube.
Specifically, ``Kinetics + COIN'' denotes that we first trained a TSN model~\cite{TSN2016ECCV} for the step recognition task on COIN, where the backbone of TSN model was pre-trained on the Kinetics dataset. 
Then we employed the backbone to the SSN model~\cite{DBLP:conf/iccv/ZhaoXWWTL17} for the Breakfast, JIGSAWS or UNLV-diving dataset.
Besides, we include the results based on two state-of-the-art methods (\textit{i.e.,} BMN~\cite{BMN} and BSN~\cite{BSN}) to see
if the improvements are subtle or significant.

\linespread{1.2}
\begin{table}[t]\small
\caption{Comparisons of the cross dataset transfer accuracy (\%) on three datasets based on the RGB modality. * and ** denote the model used ``Kinetics'' or ``Kinetics+COIN'' strategy as we introduce in the text.
}
\centering
\begin{tabular}{l | c c c  c c | c }
\toprule[1.5pt]
 \multicolumn{7}{c}{Breakfast Dataset~\cite{DBLP:conf/cvpr/KuehneAS14} (32 similar tasks in COIN), mAP @ $\alpha$} \\
\toprule[1.5pt]
Threshold & 0.1 &  0.2 & 0.3
& 0.4 & 0.5  & Average\\
\hline
BSN~\cite{BSN}  &25.01 &23.96&21.30&19.16&16.46&21.17\\
\hline
BMN~\cite{BMN} &24.40&23.19&21.38&19.49&16.88&21.07 \\
\hline
SSN* & 25.78 & 23.38 & 19.85 & 16.77 & 13.43 & 19.84 \\
SSN** & 27.47 & 24.98 & 21.14 & 18.29 & 14.94&  21.36 \\
\toprule[1.5pt]
 \multicolumn{7}{c}{JIGSAWS Dataset~\cite{gao2014jhu} (13 similar tasks in COIN), mAP @ $\alpha$} \\
\toprule[1.5pt]
Threshold & 0.1 &  0.2 & 0.3
& 0.4 & 0.5  & Average\\
\hline
BSN~\cite{BSN}  &40.58&32.25&28.08&22.84&18.35&28.42\\
\hline
BMN~\cite{BMN} &36.41&36.34&32.66&27.24&23.23&31.18 \\
\hline
SSN* &30.21&22.51&14.00&9.88&7.18&16.76 \\
SSN** & 31.01&24.08&15.79&12.38&6.30&17.91\\
\toprule[1.5pt]
 \multicolumn{7}{c}{UNLV-Diving Dataset~\cite{DBLP:conf/cvpr/ParmarM17} (0 similar task in COIN), mAP @ $\alpha$} \\
\toprule[1.5pt]
Threshold & 0.1 &  0.2 & 0.3
& 0.4 & 0.5  & Average\\
\hline
BSN~\cite{BSN}  &67.56&63.44&58.62&56.28& 52.61&59.70\\
\hline
BMN~\cite{BMN} & 75.87&73.55&67.15&62.00&54.23&66.56 \\
\hline
SSN* & 73.00 &54.07&33.26&32.44&32.44&45.04 \\
SSN** & 73.73&54.80&34.01&33.25&33.25&45.81 \\
\bottomrule[1.5pt]
\end{tabular}
\label{tab:ctt}  
\vspace{-0.3cm} 
\end{table}

\linespread{1}

In Table~\ref{tab:ctt}, 
we present the experimental results and the number of similar tasks in COIN for the three datasets.
The ``similar tasks'' here refers to the tasks with similar human-object interaction behaviors.
We observe that the model pre-trained on ``Kinetics + COIN'' achieves consistent improvements over that only pre-trained on Kinetics.
However, the gains on the UNLV-Diving dataset are very slight (\textit{e.g.,} only 0.77\% improvement on average mAP)
because of the large difference with the tasks in COIN dataset.
In comparison, the improvements on the Breakfast dataset are much more promising.
The reason is that there are 32 tasks related to food in the COIN dataset, which are similar to the tasks in the Breakfast dataset.
For the JIGSAWS dataset, it contains three surgical activities as "suturing", "knot-tying" and "needle-passing",
and there are about 13 similar tasks in the COIN dataset.
Hence, the improvements on the JIGSAWS dataset are not significant.
From the results of these three datasets, 
we further conclude that the annotation and data from COIN could lead to better results for the dataset which contains more similar tasks.
This makes sense and has also been verified in image-based datasets~\cite{DBLP:conf/nips/YosinskiCBL14}.
And also, the COIN would not hinder the performance when applied to other irrelevant tasks.

Besides, we have also attempted to evaluate a model only pre-trained on COIN.
However, when training the TSN model from scratch on the COIN, we found the performance was inferior.
This is because in the COIN dataset, the samples of each step are still limited. 
And the reason that ``Kinetics + COIN'' works better than Kinetics is that the COIN contains step annotations at finer level,
which would be more helpful for instructional video where fine-grained actions occur.

\section{Future work}
Finally, we discuss some several potential directions for future works based on our COIN dataset.

(1) \textit{Mining shared information of different steps.}
In our COIN dataset, we define different steps at task level, 
so no steps are shared between tasks. Though in different tasks, 
some steps might be similar, 
we still assign different step labels to them because of the difference in the interacted objects and the interacted ways. 
As mentioned in ~\cite{cross-task},
the shared components of different steps across takes are useful cues for analyzing instructional videos.
It is interesting to explore the shared knowledge across different tasks from two aspects: (i) Leveraging
the similarity information between tasks for step localization. (ii) Redefining the step at the dataset level to merge the similar steps in different tasks.

(2) \textit{End-to-end training.}
Since our methods attempt to refine the proposal scores during the inference stage,
others might doubt whether end-to-end training might help.
Theoretically, end-to-end training with our proposed methods would bring more improvements.
However, the bottleneck is the computational cost during implementation, 
since the frames in hundreds of proposals in a video need to be processed at the same time.
It is desirable to explore some effective ways or other methods (\textit{e.g.,} considering task and step labels simultaneously by multi-task learning) for end-to-end training in future.

(3) \textit{Semi-supervised, unsupervised or self-supervised learning for step localization.} 
Since the temporal annotation are expensive, other settings for step localization are encouraged to study based on COIN dataset besides fully-supervised learning.
For example, 
(i) semi-supervised learning setting only based on the step labels with\cite{DBLP:journals/corr/abs-1805-06875} or without\cite{richard2017action} ordering information,
(ii) unsupervised learning setting when some auxiliary information is further provided (\textit{e.g.,} narration associated with the original video can be obtained via the Automatic Speech Recognition (ASR) system~\cite{DBLP:journals/pami/AlayracBASLL18,NCE}), and
(iii) self-supervised learning setting which leverages the intrinsic long-term dependency of instructional videos (\textit{e.g.,} the recent proposed videoBert model~\cite{sun2019videobert}).
Note that though our temporal annotation can be absent for these settings during training phase,
it is still essential in evaluation period.

(4) \textit{Other tasks for instructional video analysis}. As we mentioned in Section 2.1, there are various tasks for instructional video analysis. 
Based on the existing annotation, our COIN dataset can be used to evaluate other tasks such as activity anticipation\cite{DBLP:conf/cvpr/FarhaRG18} or procedure planning\cite{DBLP:journals/corr/abs-1907-01172}. 
It can also be utilized for skill assessment\cite{EPIC-skill,USDL} and visual object grounding\cite{Huang_2018_CVPR} if the corresponding annotations are further available.

\section{Conclusions}
In this paper we have introduced COIN, 
a new large-scale dataset for comprehensive instructional video analysis.
Organized in a rich semantic taxonomy, 
the COIN dataset covers boarder domains and contains more tasks than the most existing instructional video datasets.
In order to establish a benchmark,
we have evaluated various approaches under different scenarios on the COIN.
In addition, we have explored the relationship among different steps of a specific task based on the task-consistency and ordering-dependency characteristics of instructional videos.
The experimental results have revealed the great challenges of COIN and demonstrated the effectiveness of our methods.
It has also been shown that the COIN can contribute to the step localization task for other instructional video datasets.

\section*{Acknowledgement}
This work was supported in part by the National Key Research and Development Program of China under Grant 2017YFA0700802, in part by the National Natural Science Foundation of China under Grant 61822603, Grant U1813218, Grant U1713214, and Grant 61672306, in part by the Shenzhen Fundamental Research Fund (Subject Arrangement) under Grant JCYJ20170412170602564, and in part by Tsinghua University Initiative Scientific Research Program.

The authors would like to thank Dajun Ding and Lili Zhao from Meitu Inc. for their helps on computing resources and annotation, Danyang Zhang, Yu Zheng and Xumin Yu for conducting partial experiments, Yongming Rao for valuable discussion.

\begin{appendices}
\section{A brief review of some instructional video datasets}
We briefly review some representative datasets as follow:

(1) The \textit{MPII}\cite{DBLP:conf/cvpr/RohrbachAAS12} and \textit{Breakfast}\cite{DBLP:conf/cvpr/KuehneAS14} datasets are two self-collected instructional video datasets in the early time.
As a pioneering work, Rohrbach \textit{et al.} proposed the MPII dataset which contained 44 long videos.
They provided annotations of 5,609 clips on 65 fine-grained cooking activities.
Later, Kuehne \textit{et al.} introduced the Breakfast dataset which consisted of 1,989 videos.
This dataset included 10 cooking activities (tasks) of 52 participants.
The annotation also contained temporal intervals of 48 action units (steps).
The main purpose of these two datasets is to detect the main steps and recognize their labels.

(2) The \textit{YouCook}\cite{DBLP:conf/cvpr/DasXDC13} and \textit{YouCook2}\cite{DBLP:conf/aaai/ZhouXC18} datasets were collected by downloading cooking activity videos from YouTube.
In 2013, Das \textit{et al.} introduced the YouCook dataset, which consisted of 88 videos.
Each video was annotated with at least three sentences by the participants in MTurk.
More recently in 2018, Zhou \textit{et al.} collected the YouCook2 dataset of 2,000 videos, 
which were labelled by temporal intervals of different steps and their captions on the recipe.
As an extension work, they further provided the object level annotation~\cite{DBLP:conf/bmvc/ZhouLC18}.
Both the YouCook and YouCook2 datasets can be used for the video caption tasks,
and the YouCook2 further facilitated the research for procedure segmentation~\cite{DBLP:conf/aaai/ZhouXC18} and video object grounding~\cite{DBLP:conf/bmvc/ZhouLC18,Huang_2018_CVPR}.

(3) The \textit{``5 tasks''}\cite{DBLP:conf/cvpr/AlayracBASLL16} and \textit{HowTo100M}\cite{miech19howto100m} datasets were collected for unsupervised learning from narrations.
The ``5 tasks'' dataset~\cite{DBLP:conf/cvpr/AlayracBASLL16} consisted of 150 videos on 5 tasks (changing tire, performing CPR, repoting plant, making coffee and jumping cars).
Each video was associated with a sequence of natural description text, which was generated from the corresponding audio modality.
The goal of this dataset was to automatically discover the main steps of a certain task and locate them in the videos in an unsupervised setting.
More recently, Miech \textit{et al.}\cite{miech19howto100m} introduced another large-scale dataset HowTo100M, 
which contained 136 million clips sourced from 1.238 million instructional videos.
These video clips were paired with a list of text chunks based on their corresponding time intervals.
With its large-scale data, 
the HowTo100M greatly promoted the development of pre-trained text-video joint embedding models for various vision-languages tasks.
However, as the authors mentioned, since their annotations were automatically generated from the narration,
there might be various of incoherent samples.
For example, the narration was unrelated to the video, 
or describing something before or after it happens in video.
Hence, our manually labelled COIN is complementary to HowTo100M from this aspect.

(4)
The \textit{EPIC-Skills}\cite{EPIC-skill} and \textit{BEST}\cite{BEST} datasets were constructed for skill determination, 
which referred to assess the behaviour of a subject to accomplish a task.
For both datasets, the AMT workers were asked to watch the videos in a pair-wise manner, 
and selected the video which contained more skill.
The BEST dataset also provided the initial opinion of the annotators over the video as ``Beginner'', ``Intermediate'' or ``Expert''.

(5)
The \textit{CrossTask}\cite{cross-task} dataset contained 4.7k videos of 83 tasks, and the goal was to localize the steps in the video by weakly supervised learning (\textit{i.e.,} instructional narrations and an ordered list of steps).
Specifically, this dataset was proposed to assess the shared components on different steps across different tasks.
For example, the task ``pour egg'' would be benefit from other tasks involving ``pour'' and ``egg''.

\linespread{1.3}
\begin{table}[t]
\caption{Comparisons of the annotation time cost under two modes. FM indicates the new developed frame mode, and VM represents the conventional video mode.}
\centering
\begin{tabular}{ r  c  r r }
\toprule[1.5pt]
Task & samples  &  FM & VM\\
\hline
Assemble Bed  &  6  & 6:55 & 23:30 \\
Boil Noodles  &  5   & 3:50 & 18:15 \\
Lubricate A Lock   & 2   & 1:23 & 5:29 \\
Make French Fries &  6   & 5:57 &20:24 \\
Change Mobile Phone Battery &  2   & 2:23 & 7:35 \\
Replace A Bulb &  2   & 1:30  & 6:40\\
Plant A Tree &  2   & 1:45  & 6:37\\
\hline
Total & 25 &  23:43 & 88:30  \\
\bottomrule[1.5pt]
\end{tabular}
\label{tab:at}
\end{table}

\linespread{1}

\section{Annotation Time Cost Analysis}
In section 3.2, we have introduced a toolbox for annotating COIN dataset.
The toolbox has two modes:  frame mode and video mode.
The frame mode is new developed for efficient annotation, while the video mode is frequently used in previous works~\cite{DBLP:conf/iccv/KrishnaHRFN17}.
We have evaluated the annotation time on a small set of COIN, which contains 25 videos of 7 tasks.
Table \ref{tab:at} shows the comparison of annotation time under two different modes.
We observe that the annotation time under the frame mode is only 26.8\% of that under the video mode, 
which shows the advantages of our toolbox.

\begin{figure*}[!t]
\includegraphics[width = \linewidth]{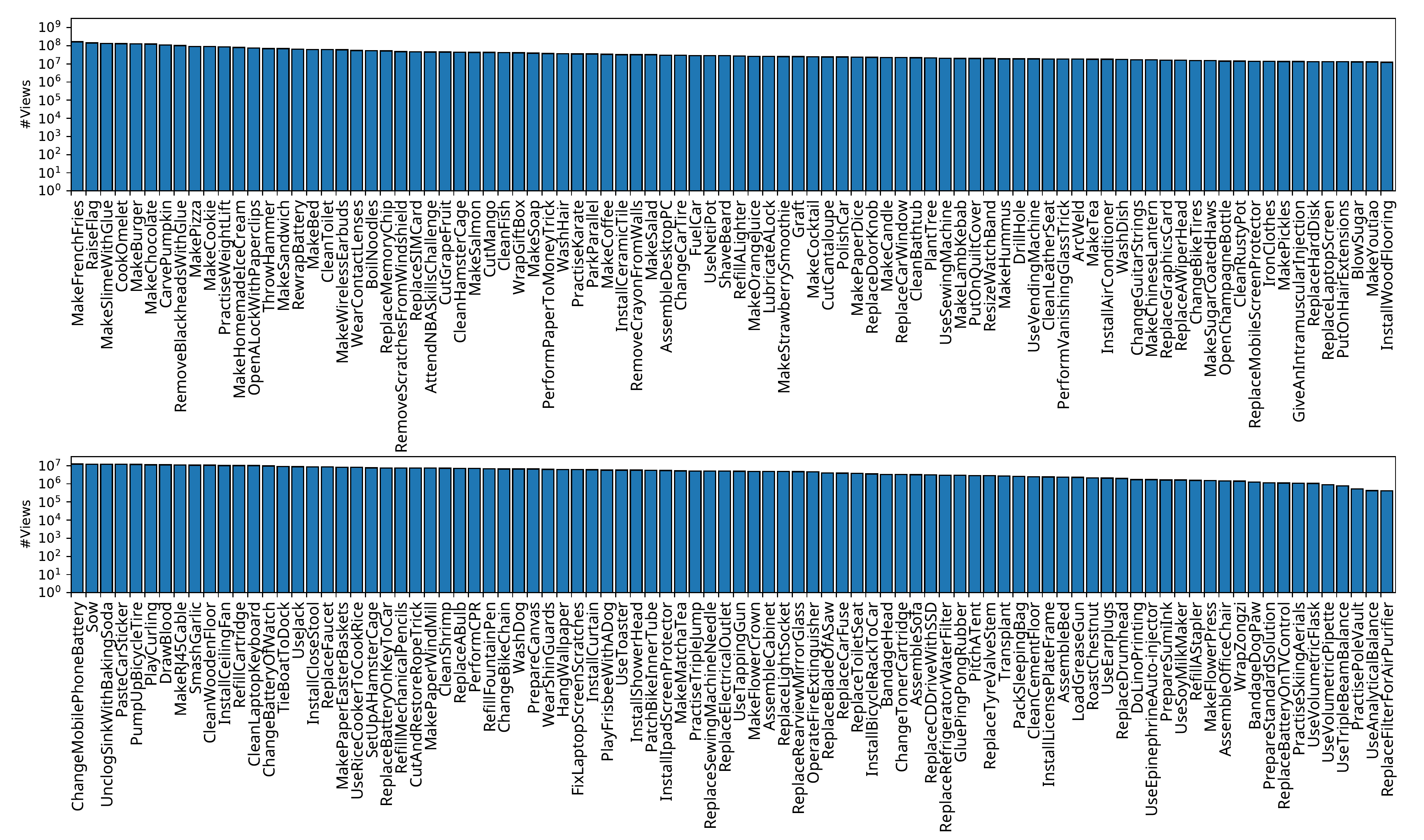}
\caption{The browse time distributions of the selected 180 tasks on YouTube.}
\label{fig:bt}
\end{figure*}

\section{Browse Times Analysis}
In order to justify that the selected tasks meet the need of website viewers, 
we display the number of browse times across 180 tasks in Fig. \ref{fig:bt}. 
We searched ``How to'' + name of 180 tasks, e.g., ``How to Assemble Sofa'', on YouTube respectively. 
Then we summed up the browse times of the videos appearing in the first pages (about 20 videos) to get the final results.
``Make French Fries'' is the most-viewed task, which has been browsed $1.7 \times 10^8$ times.
And the browse times per task are $2.3 \times 10^7$ on average.
These results demonstrate the selected tasks of our COIN dataset satisfy the need of website viewers,
and also reveal the practical value of instructional video analysis.

\begin{figure}[!t]
\centering
\includegraphics[width = \linewidth]{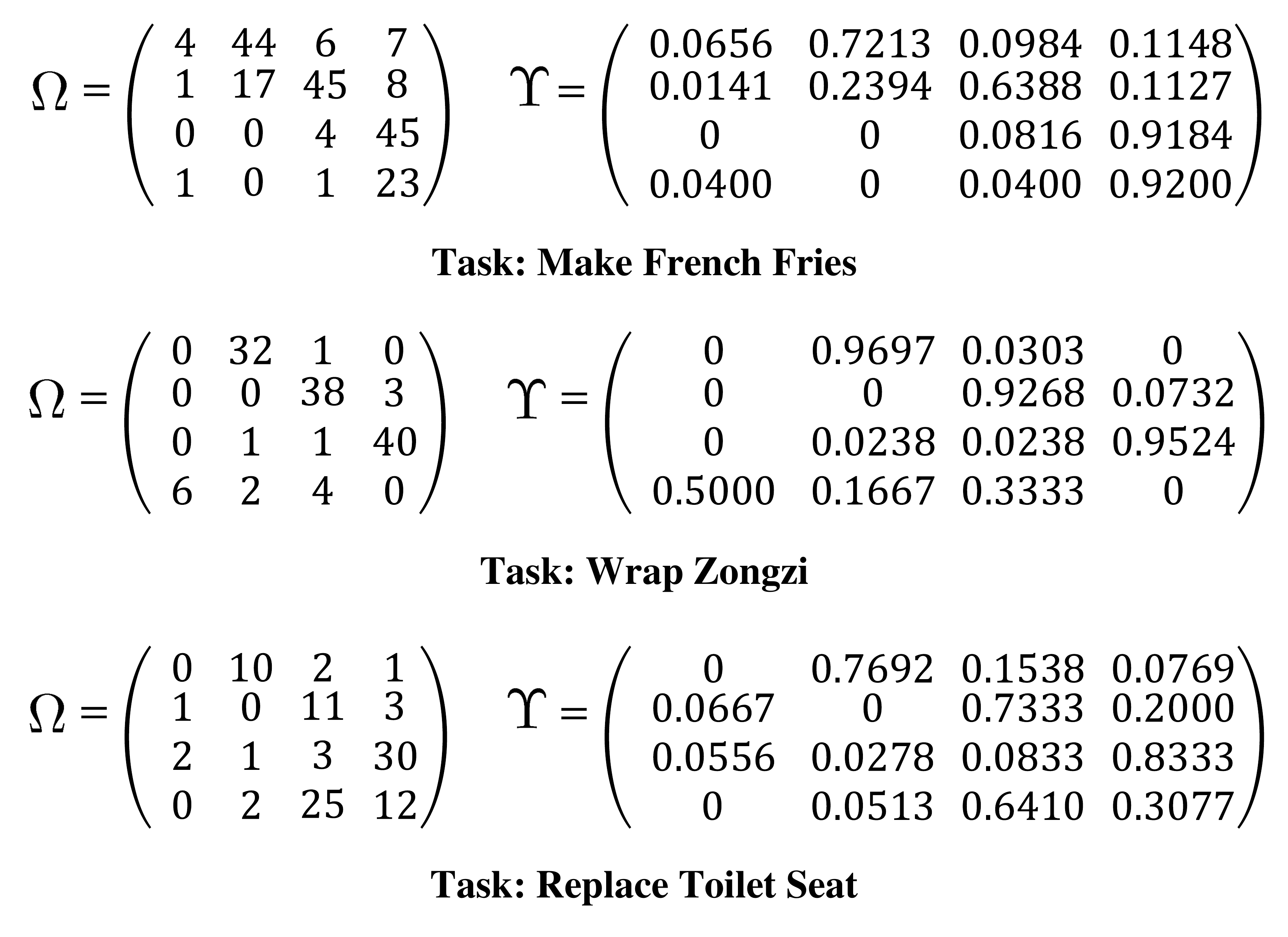}
\caption{
Visualization of auxiliary matrix $\Omega$ and transition matrix $\Upsilon$ of three tasks. Recall that the auxiliary matrix $\Omega$ is constructed by counting the occurrence time of the step $i$ after step $j$ based on the all ordered step lists appearing in the training set as
$
    \Omega_{ij} = \#\textit{(step j occurs after step i).}
$
We normalize each row of $\Omega$ to obtain the transition matrix $\Upsilon$, in which the element $\Upsilon_{ij}$ of the transition matrix denotes the probability of step $i$ occurs after step $j$ as 
$
   \Upsilon_{ij} = p(S^l = j | S^{l-1} = i), \quad l = 2,3,..,L.
$. The step lists of these tasks are: $\{$cut potato into strips,
soak them in water, dry strips, put in the oil to fry$\}$, $\{$cone the leaves, add ingredients into cone, fold the leaves, tie the zongzi tightly$\}$ and $\{$take out the screws, remove the old toilet seat, install the new toilet seat, screw on the screws$\}$.
}
\label{fig:trans-mat}
\end{figure}

\section{Visualization of matrices}
In section 4.2, we introduce the transition matrix $\Upsilon$ and the corresponding auxiliary matrix $\Omega$ for the OD method.
Here we visualize the matrices of three different tasks in Fig.~\ref{fig:trans-mat},
where the definitions of $\Upsilon$ and $\Omega$ are detailed in caption of the figure.

\section{Sample distribution of COIN}
\begin{figure*}[!t]
\includegraphics[width = \linewidth]{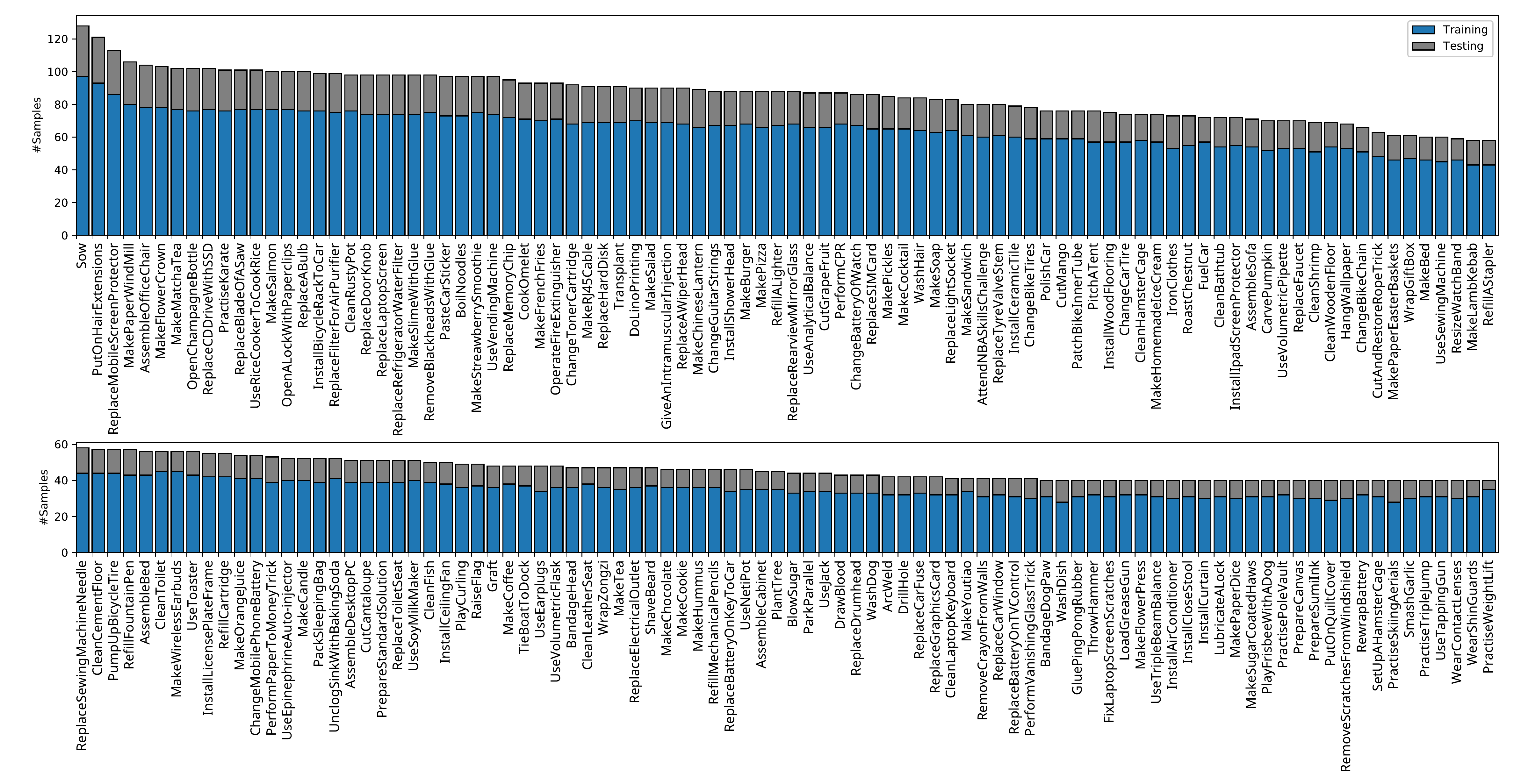}
\caption{
The sample distributions of all the tasks in COIN.
The blue bars and the grey bars indicate the number of training and testing videos in each class respectively. 
}
\label{fig:sam_dis}
\end{figure*}

We present the sample distributions of all the tasks in COIN in Fig. \ref{fig:sam_dis}.
To alleviate the effect of long tails,
we make sure that there are more than 39 videos for each task.

\linespread{1.3}
\begin{table*}[t]

\caption{Clarification of different tasks evaluated on the COIN.} 
\small
\centering 
\begin{tabular}{p{90pt}|p{130pt}|p{110pt}|p{110pt}}
\toprule[1.5pt]
Task & Goal & Evaluation Metrics& Evaluated Methods \\
\hline
step localization & localize the step boundary and predict the step label & mAP/mAR (interval level) & R-C3D\cite{DBLP:conf/iccv/XuDS17}, SSN\cite{DBLP:conf/iccv/ZhaoXWWTL17}, TC(ours), OD(ours)\\
\hline
action segmentation &  assign each frame a step label &  accuracy (frame level) & Action-Sets\cite{richard2017action}, NN-Viterbi\cite{DBLP:journals/corr/abs-1805-06875}, TCFPN-ISBA\cite{DBLP:journals/corr/abs-1803-10699} \\ 
\hline
proposal localization & localize the step boundary  & mAP/mAR (interval level) & TAG\cite{DBLP:conf/iccv/ZhaoXWWTL17} \\
\hline
task recognition &  recognize the \textit{task} label  & accuracy (video level ) & TSN\cite{TSN2016ECCV} \\
\hline
step recognition &  recognize the \textit{step} label given the step boundary & accuracy (interval level) & TSN\cite{TSN2016ECCV}\\
\bottomrule[1.5pt]
\end{tabular}
\label{tab:setting}
\end{table*}
\linespread{1}

\section{Wordles}
\begin{figure}[!t]
\includegraphics[width = \linewidth]{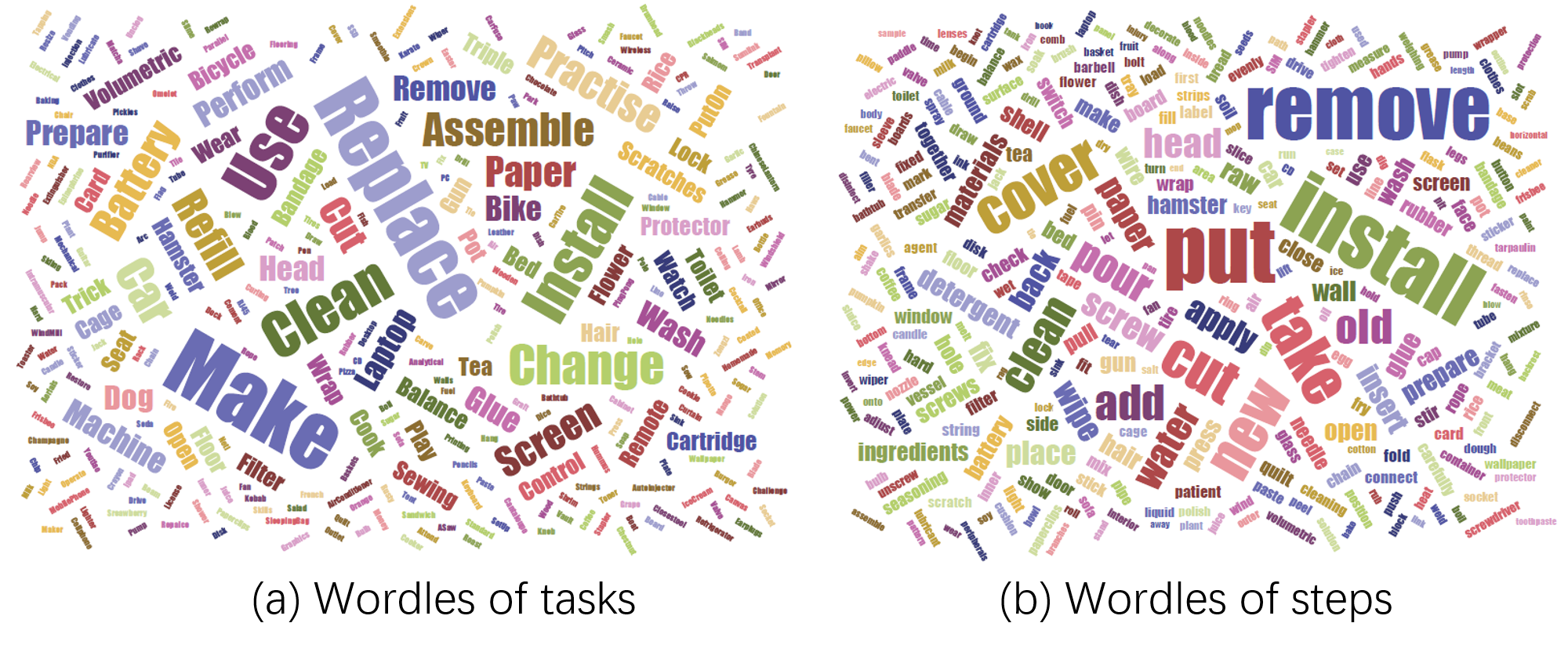}
\caption{Wordles of the annotations of COIN. The figure of tasks (the second-level tags) is shown on the left, while the figure of steps (the third-level tags) is presented on the right.
}
\label{fig:word}
\end{figure}

We show the wordles of the annotation of COIN in Fig. \ref{fig:word}.
There are 4.84 words per phrase for each step.

\section{Clarification of different tasks evaluated on the COIN}
In the experiment section, we evaluate five tasks (i.e., step localization, action segmentation, procedure localization, task recognition and step recognition) on the COIN dataset. For clarification, we present the goal, metric, and evaluated methods for each task in Table \ref{tab:setting}.

\linespread{1.2}
\begin{table*}[t]\normalsize
\setlength{\tabcolsep}{9.5pt}
  \centering
  \caption{Study of the attenuation coefficient $\gamma$ on the COIN dataset.}
  \begin{tabular}{c|ccccc|ccccc}
    \hline
    Metrics & \multicolumn{5}{c}{mAP @ $\alpha$} & \multicolumn{5}{|c}{mAR @ $\alpha$} \\
    & 0.1 & 0.2 & 0.3 & 0.4 & 0.5 & 0.1 & 0.2 & 0.3 & 0.4 & 0.5 \\
    \hline
    0 &     19.95 & 16.64 & 14.12 & 11.69 & 9.30 &  52.00 & 45.60 & 39.74 & 34.26 & 28.55   \\
    $e^{-3}$ &  20.12 & 16.77 & 14.23 & 11.79 & 9.38 &  53.38 & 46.79 & 40.69 & 34.98 & 29.14   \\
    $e^{-2}$ &  20.15 & 16.79 & 14.24 & 11.74 & 9.33 &  54.05 & 47.31 & 40.99 & 35.11 & 29.17   \\
    $e^{-1}$ &  20.10 & 16.76 & 14.19 & 11.71 & 9.30 &  54.48 & 47.51 & 41.09 & 35.36 & 29.36   \\
    \hline
  \end{tabular}
  \label{tab:gamma}
\end{table*}
\linespread{1}

\linespread{1.2}
\begin{table*}[t]\normalsize
\setlength{\tabcolsep}{9.5pt}
  \centering
  \caption{Study of the number of time slots M on the COIN dataset.}
  \begin{tabular}{c|ccccc|ccccc}
    \hline
    Metrics & \multicolumn{5}{c}{mAP @ $\alpha$} & \multicolumn{5}{|c}{mAR @ $\alpha$} \\
    & 0.1 & 0.2 & 0.3 & 0.4 & 0.5 & 0.1 & 0.2 & 0.3 & 0.4 & 0.5 \\
    \hline
    50 &  19.87 & 15.97 & 12.95 & 10.22 & 7.98 &  51.19 & 43.73 & 37.33 & 31.84 & 26.58 \\
    100 & 20.39 & 16.35 & 13.20 & 10.38 & 8.13 &  51.51 & 43.87 & 37.37 & 31.81 & 26.64 \\
    150 & 20.55 & 16.37 & 13.20 & 10.36 & 8.10 &  51.67 & 43.92 & 37.45 & 31.86 & 26.65 \\
    200 & 20.58 & 16.42 & 13.20 & 10.41 & 8.15 &  51.84 & 44.04 & 37.38 & 31.74 & 26.59 \\
    \hline
  \end{tabular}
  \label{tab:nb_sls}
\end{table*}
\linespread{1}

\section{Experiments on the hyper-parameters}
Table \ref{tab:gamma} presents the results of $\gamma$ (introduced in section 4.1).
On one hand, $\gamma$ cannot be too large as it is an attenuation coefficient in the TC method.
On the other hand, it cannot be too small, because if the selected task is wrong (sometimes the
scores of two different tasks are approached), the steps in
other tasks should also be taken into consideration.
Experiments show that $e^{-2}$ is more effective for $\gamma$ in our case.

Table \ref{tab:nb_sls} presents the evalutation results of the time slot M (introduced in section 4.2).
It can be seen that finer-grained division with larger M can lead to better performance.
In this paper, we use $M=100$ to make a good trade-off between the effectiveness and efficiency as larger M would also bring more computational cost.

\begin{figure}[!t]
  \centering
  \includegraphics[width=\linewidth]{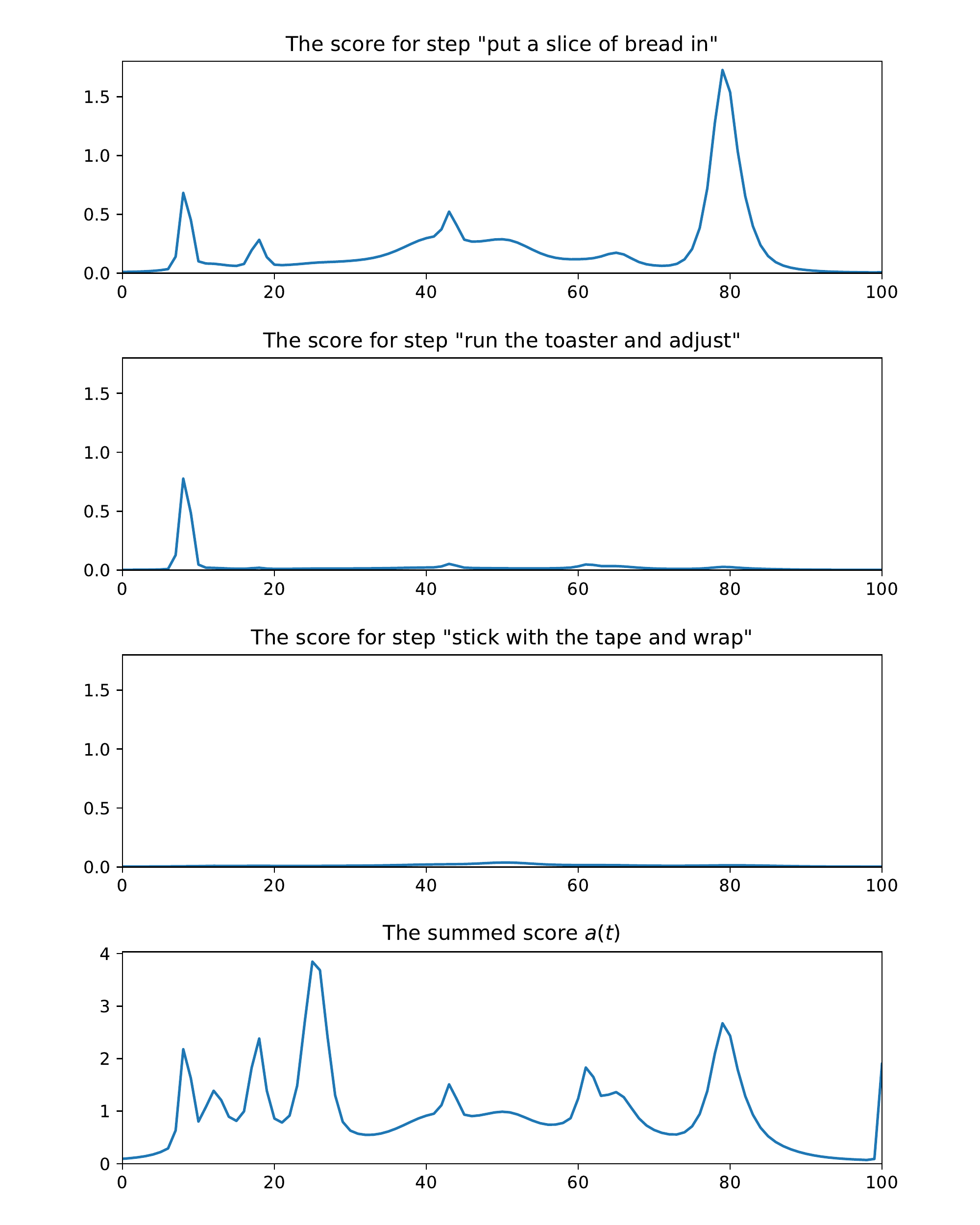}
  \caption{Visualization of different step scores in $\textbf{f}(t)$ and the summed score a(t). The video belongs to the task ``Use Toaster''. In the above plots, the ``put a slice of bread in'' and ``run the toaster and adjust'' are two steps of this task while ``stick with the tap and wrap'' is not.}
  \label{fig:vslzt_of_f_t}
\end{figure}

\section{Visualization of \textbf{f}(t)}
In section 4.2,
\textbf{f}(t) denotes the scores of different steps, which implies their possibilities occuring at the time-stamp t.
We show the visualization of several steps of \textbf{f}(t) when evaluating OD method in Fig. \ref{fig:vslzt_of_f_t}.
The summed score, which is denoted as a(t) in our paper, is also presented.

\section{Analysis on the Watershed Algorithm}
In section 4.2, we employ the watershed algorithm~\cite{DBLP:conf/iccv/ZhaoXWWTL17,DBLP:journals/fuin/RoerdinkM00} to obtain a sequence of segments during the period of grouping proposals. Particularly, we iterate the thresholds of the actionness score from high value ($0.95 \max_t a(t)$) to low value ($0.05 \max_t a(t)$) until the termination criterion is met and consider the intervals where $a(t)$ is larger than the theshold as the action segments. 

Here we study two types of termination criteria. The first one is to do the iteration until the average temporal gap between different segments gets smaller than a specific value $\theta_{G}$. Another one is to check if the average temporal length of the segments is greater than a specific value $\theta_{L}$.
Table \ref{tab:tmnt_crtr_of_wtsh_mth}
shows the experimental results of these two strategies.
And in our main paper, we use ``$\theta_{G}$=6'' as the termination criterion of the watershed algorithm.

\linespread{1.3}
\begin{table*}[tb] \normalsize
  \centering
  \caption{Study of two hyper-parameters in the watershed algorithm. 
$\theta_{G}$ denotes the average temporal gap between different segments, and $\theta_{L}$ denotes the average temporal length of different segments.
We used the time slots as the unit for these two hyperparameters in this table.}
  \begin{tabular}{c|c|ccccc|ccccc}
    \hline      
    &  &  \multicolumn{5}{c|}{mAP @ $\alpha$} & \multicolumn{5}{c}{mAR @ $\alpha$} \\
    Criteria &  Parameters &  0.1 & 0.2 & 0.3 & 0.4 & 0.5 & 0.1 & 0.2 & 0.3 & 0.4 & 0.5 \\
    \hline
     &  2 & 19.75 & 15.80 & 12.72 & 9.95 &  7.79 &  51.06 & 43.48 & 36.96 & 31.44 & 26.25 \\
    & 4 & 20.22 & 16.20 & 13.07 & 10.26 & 8.04 &  51.25 & 43.66 & 37.16 & 31.64 & 26.45 \\
  $\theta_{G}$ &  6 & 20.39 & 16.35 & 13.20 & 10.38 & 8.13 &  51.51 & 43.87 & 37.37 & 31.81 & 26.64 \\
    & 8 & 20.41 & 16.36 & 13.21 & 10.39 & 8.13 &  51.60 & 43.97 & 37.40 & 31.81 & 26.66 \\
    & 10 &  20.39 & 16.35 & 13.23 & 10.34 & 8.11 &  51.62 & 43.99 & 37.37 & 31.71 & 26.60 \\
    \hline
    & 10 &  20.07 & 16.06 & 12.92 & 10.13 & 7.93 &  51.70 & 44.06 & 37.42 & 31.87 & 26.60 \\
  $\theta_{L}$  & 15 &  19.98 & 15.99 & 12.87 & 10.07 & 7.89 &  51.58 & 43.95 & 37.33 & 31.77 & 26.54 \\
    & 20 &  19.96 & 15.99 & 12.88 & 10.09 & 7.91 &  51.41 & 43.83 & 37.30 & 31.76 & 26.49 \\
    \hline
  \end{tabular}
  \label{tab:tmnt_crtr_of_wtsh_mth}
\end{table*}

\linespread{1.3}
\begin{table*}[t]\normalsize
\setlength{\tabcolsep}{9.5pt}
  \centering
  \caption{Study of the OD approach on different basic models. The results are reported based on the RGB modality in the COIN dataset. Since the BMN~\cite{BMN} and BSN\cite{BSN} are originally designed for temporal action proposal generation, 
we process the proposals generated by these two methods with the classifier of SSN~\cite{DBLP:conf/iccv/ZhaoXWWTL17} to produce the final results.}
  \begin{tabular}{c|ccccc|ccccc}
    \hline
    Metrics & \multicolumn{5}{c}{mAP @ $\alpha$} & \multicolumn{5}{|c}{mAR @ $\alpha$} \\
    & 0.1 & 0.2 & 0.3 & 0.4 & 0.5 & 0.1 & 0.2 & 0.3 & 0.4 & 0.5 \\
    \hline
    BMN~\cite{BMN} &    18.60 & 16.76 & 14.78 & 12.40 & 10.38 & 48.70 & 45.71 & 42.45 & 38.70 & 34.57 \\
    BMN+TC &  19.27 & 17.17 & 15.09 & 12.60 & 10.59 & 47.87 & 44.64 & 41.61 & 37.92 & 34.07 \\
    BMN+OD &  19.68 & 17.44 & 15.27 & 12.90 & 10.75 & 49.70 & 46.00 & 42.43 & 38.78 & 34.64 \\
    \hline
    BSN~\cite{BSN} &    18.91 & 16.84 & 14.49 & 12.26 & 10.00 & 46.97 & 43.87 & 39.78 & 35.77 & 31.61 \\
    BSN+TC &  19.96 & 17.54 & 15.00 & 12.68 & 10.35 & 47.16 & 44.16 & 40.12 & 36.40 & 32.45 \\
    BSN+OD &  20.46 & 17.82 & 15.20 & 12.85 & 10.34 & 48.76 & 44.88 & 40.86 & 36.95 & 32.54 \\
    \hline
R-C3D\cite{DBLP:conf/iccv/XuDS17} &   9.85 & 7.78 & 5.80 & 4.23 & 2.82  & 36.82 & 31.55 & 26.56 & 21.42 & 17.07 \\    
R-C3D+TC &  10.32 & 8.25 & 6.20 & 4.54 & 3.08 & 39.25 & 34.22 & 29.09 & 23.71 & 19.24 \\    
    R-C3D+OD &  10.08 & 8.01 &  5.98 &  4.36 &  2.93 &  37.37 & 31.84 & 26.81 & 21.74 & 14.43     
    \\
    \hline
  \end{tabular}
  \label{tab:bac_mod}
\end{table*}
\linespread{1}

\linespread{1.3}
\begin{table*}[!t]\normalsize
\setlength{\tabcolsep}{9.5pt}
  \centering
  \caption{Study of different generated distributions for calculating scores in OD method on the COIN dataset.}
  \begin{tabular}{c|ccccc|ccccc}
    \hline
    Metrics & \multicolumn{5}{c}{mAP @ $\alpha$} & \multicolumn{5}{|c}{mAR @ $\alpha$} \\
    Distributions & 0.1 & 0.2 & 0.3 & 0.4 & 0.5 & 0.1 & 0.2 & 0.3 & 0.4 & 0.5 \\
    \hline
    Gaussian, $\beta = 0.5$ & 20.07 & 15.90 & 12.78 & 10.00 & 7.81 &  51.05 & 43.30 & 36.81 & 31.28 & 26.18 \\
    Gaussian, $\beta = 1$ & 20.39 & 16.35 & 13.20 & 10.38 & 8.13 &  51.51 & 43.87 & 37.37 & 31.81 & 26.64 \\
    Gaussian, $\beta = 2$ & 20.13 & 16.20 & 13.13 & 10.34 & 8.08 &  51.40 & 43.91 & 37.46 & 31.93 & 26.65 \\
    Gaussian, $\beta = 5$ & 19.81 & 15.97 & 12.92 & 10.18 & 7.98 &  51.19 & 43.85 & 37.34 & 31.78 & 26.58 \\
\hline
    Triangle &  19.77 & 15.66 & 12.55 & 9.78 &  7.66 &  50.99 & 43.23 & 36.64 & 31.08 & 26.03 \\
    \hline
  \end{tabular}
  \label{tab:dtrbt_fun}
\end{table*}
\linespread{1}

\section{Detailed Results on the OD method and other action detectors}
In our main paper, Table 6 presents the results on different generated distributions $\textbf{f}^n(t)$ and fusion methods for OD method,
and Table 8 shows the results on more basic action detectin models.
Here we further present more detailed results on these issues in Table \ref{tab:bac_mod}, Table \ref{tab:dtrbt_fun} and Table \ref{tab:fus_eqt}.

\linespread{1.3}
\begin{table*}[t]\normalsize
\setlength{\tabcolsep}{9.5pt}
  \centering
  \caption{Study of different methods to refine the segment scores on the COIN dataset. We set the hyper-parameters $\lambda_1, \lambda_2$ as $\lambda_1 = \lambda_2 = 0.5$. All the fusion methods are performed on the element-wise level for the scores $\textbf{s}_1$ and $\textbf{s}_2$.}
  \begin{tabular}{c|ccccc|ccccc}
    \hline
     & \multicolumn{5}{c}{mAP @ $\alpha$} & \multicolumn{5}{|c}{mAR @ $\alpha$} \\
Fusion methods  & 0.1 & 0.2 & 0.3 & 0.4 & 0.5 & 0.1 & 0.2 & 0.3 & 0.4 & 0.5 \\
    \hline
    $\lambda_1 \textbf{s}_1 + \lambda_2 \textbf{s}_2$ & 20.30 & 16.28 & 13.20 & 10.37 & 08.12 & 51.51 & 43.93 & 37.47 & 31.88 & 26.69 \\
    $\sqrt{\lambda_1 \textbf{s}_1^2 + \lambda_2 \textbf{s}_2^2}$ &  20.09 & 16.18 & 13.12 & 10.32 & 08.06 & 51.42 & 43.98 & 37.54 & 31.94 & 26.64 \\
    $\textbf{s}_1^{\lambda_1} \textbf{s}_2^{\lambda_2}$ & 20.36 & 16.32 & 13.19 & 10.36 & 08.11 & 51.57 & 43.89 & 37.37 & 31.81 & 26.64 \\
    max-pool$(\textbf{s}_1, \textbf{s}_2)$ &        19.84 & 16.01 & 12.97 & 10.21 & 07.98 & 51.41 & 44.01 & 37.51 & 31.97 & 26.70 \\
    \hline
  \end{tabular}
  \label{tab:fus_eqt}
\end{table*}
\linespread{1}

\begin{figure*}[!t]
\includegraphics[width = \linewidth]{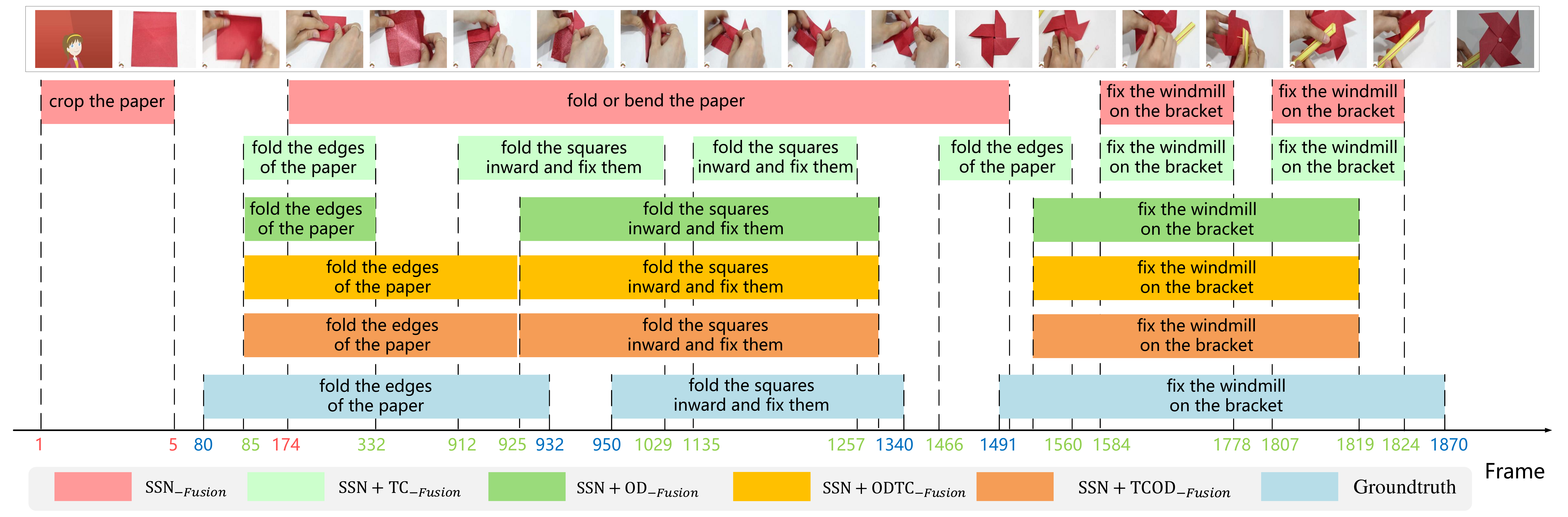}
\caption{Visualization of step localization results. 
The video is associated with the task ``make paper windmill''.
}
\label{fig:vis_sl}
\vspace{-0.1cm}
\end{figure*}

\begin{figure}[!t]
\centering
\includegraphics[width = \linewidth]{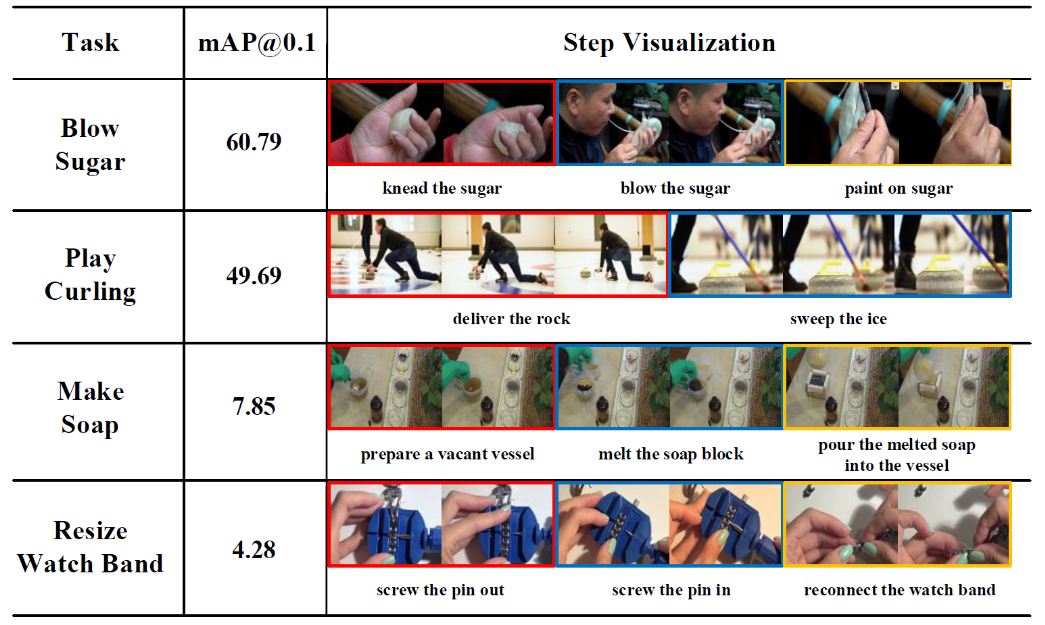}
\caption{
Comparisons of the step localization accuracy (\%) of different tasks. 
We report the results obtained by SSN+TC$_{-Fusion}$ with $\alpha$ = 0.1.
}
\label{fig:different_task}
\end{figure}

\section{Visualization Results of Different Tasks}
In section 5.2, we have compared the performance across different domains.
Fig. \ref{fig:different_task} further shows some examples from 4 different tasks as ``blow sugar'', ``play curling'', ``make soap'' and ``resize watch band''.
They belong to the domain ``sports'',
``leisure \& performance'', ``gadgets'' and ``science and craft'', which are the two of the easiest domains and the two of the hardest domains.
For ``blow sugar'' and ``play curling'',
different steps vary a lot in appearance, thus it is easier to localize them in videos.
For ``make soap'' and ``resize watch band'',
various steps tend to occur in similar scenes, hence the mAP accuracy of these tasks are inferior.

Besides Fig. 10 in our main paper, we show one more visualized step localization result in Fig. \ref{fig:vis_sl}. 
The video is associated with the task ``make paper windmill'' and the results further demonstrate the effectiveness of our proposed TC and OD methods.

\linespread{1.22}
\begin{table*}[!t]\normalsize
\caption{
Statistical analysis on the ordering characteristics of the 180 tasks in COIN dataset.
}
\setlength{\tabcolsep}{9.5pt}
\centering
\footnotesize
\begin{tabular}{ccc|ccc|ccc|ccc}
\toprule[1.5pt]
Task&MSS&OCE&
Task&MSS&OCE&
Task&MSS&OCE&
Task&MSS&OCE
\\
\hline
0&0.3937&0.2948&45&0.1167&0.1908&90&0.1429&0.1605&135&0.4078&0.1038
\\1&0.1443&0.2294&46&0.3643&0.2049&91&0.2000&0.3961&136&0.4800&0.0918
\\2&0.6020&0.3415&47&0.1327&0.1343&92&0.4580&0.2611&137&0.3502&0.0423
\\3&0.5493&0.4122&48&0.2875&0.2297&93&0.4225&0.2286&138&0.2592&0.0216
\\4&0.2889&0.1982&49&0.1902&0.4522&94&0.3494&0.1235&139&0.3590&0.1095
\\5&0.4000&0.2258&50&0.3152&0.5097&95&0.2629&0.1646&140&0.2211&0.0345
\\6&0.2333&0.1415&51&0.1957&0.3180&96&0.1667&0.1038&141&0.0465&0.1151
\\7&0.5884&0.1018&52&0.2787&0.0979&97&0.4679&0.1235&142&0.0690&0.0122
\\8&0.2907&0.0244&53&0.4495&0.1640&98&0.3967&0.1566&143&0.2500&0.0441
\\9&0.1667&0.2029&54&0.4079&0.0592&99&0.4875&0.2517&144&0.3333&0.0000
\\10&0.3840&0.0837&55&0.2174&0.0847&100&0.4291&0.3103&145&0.1181&0.0000
\\11&0.3618&0.3733&56&0.4695&0.3254&101&0.4676&0.0726&146&0.0974&0.0499
\\12&0.3833&0.1097&57&0.5659&0.0220&102&0.0885&0.4320&147&0.3578&0.2795
\\13&0.2135&0.2342&58&0.0272&0.0110&103&0.2146&0.4725&148&0.1429&0.0539
\\14&0.4857&0.1212&59&0.2195&0.0448&104&0.4200&0.3225&149&0.3063&0.1628
\\15&0.1860&0.0000&60&0.2522&0.2617&105&0.0850&0.0244&150&0.3699&0.1120
\\16&0.5403&0.2246&61&0.1282&0.0685&106&0.2903&0.1545&151&0.3313&0.1860
\\17&0.3267&0.2540&62&0.4062&0.1143&107&0.0000&0.0189&152&0.1389&0.0828
\\18&0.0788&0.0531&63&0.1571&0.3838&108&0.0252&0.2010&153&0.4362&0.1868
\\19&0.4939&0.0909&64&0.0638&0.1706&109&0.1557&0.0844&154&0.2625&0.6657
\\20&0.4408&0.3092&65&0.3562&0.1468&110&0.1591&0.2353&155&0.3000&0.3750
\\21&0.3162&0.1345&66&0.2449&0.0449&111&0.2694&0.0821&156&0.4674&0.1917
\\22&0.4640&0.3878&67&0.6343&0.0746&112&0.5632&0.5025&157&0.0353&0.1667
\\23&0.3949&0.2627&68&0.5443&0.3366&113&0.4079&0.2310&158&0.3264&0.2707
\\24&0.4964&0.2928&69&0.3583&0.1928&114&0.1833&0.0932&159&0.3489&0.1853
\\25&0.4861&0.2605&70&0.1939&0.2414&115&0.2417&0.6190&160&0.0750&0.6606
\\26&0.5544&0.3462&71&0.1288&0.0417&116&0.0583&0.6424&161&0.4250&0.0622
\\27&0.6398&0.2928&72&0.5083&0.0484&117&0.0000&0.1413&162&0.2692&0.1337
\\28&0.4247&0.1257&73&0.4907&0.4682&118&0.2843&0.1579&163&0.1100&0.2462
\\29&0.0000&0.5493&74&0.2562&0.1250&119&0.0000&0.3103&164&0.1806&0.0787
\\30&0.4816&0.2778&75&0.5228&0.5254&120&0.2561&0.1393&165&0.0238&0.4991
\\31&0.2183&0.3175&76&0.0990&0.0455&121&0.2150&0.0542&166&0.1633&0.1123
\\32&0.4000&0.1850&77&0.2500&0.3972&122&0.0716&0.3968&167&0.3580&0.2754
\\33&0.3132&0.2252&78&0.4550&0.3864&123&0.1462&0.0818&168&0.1980&0.0123
\\34&0.5450&0.2705&79&0.1894&0.0655&124&0.2311&0.1841&169&0.2627&0.0990
\\35&0.1905&0.1324&80&0.4269&0.3316&125&0.0230&0.0116&170&0.3800&0.1061
\\36&0.4167&0.2753&81&0.5634&0.5474&126&0.1455&0.0621&171&0.1905&0.1373
\\37&0.1250&0.1788&82&0.3976&0.1103&127&0.1750&0.2632&172&0.0244&0.4203
\\38&0.0250&0.6638&83&0.1801&0.0288&128&0.1900&0.0638&173&0.2448&0.3026
\\39&0.6998&0.4741&84&0.2804&0.4672&129&0.5907&0.1401&174&0.0000&0.1531
\\40&0.2553&0.1732&85&0.3563&0.2337&130&0.2079&0.0422&175&0.5068&0.5137
\\41&0.3810&0.2424&86&0.4485&0.0302&131&0.4647&0.0671&176&0.1917&0.1880
\\42&0.0922&0.3520&87&0.0926&0.0581&132&0.3095&0.1240&177&0.2833&0.3628
\\43&0.5540&0.1977&88&0.0863&0.0769&133&0.3388&0.1801&178&0.2875&0.6123
\\44&0.3095&0.0444&89&0.2699&0.1824&134&0.2442&0.1447&179&0.4139&0.0719
\\
\bottomrule[1.5pt]
\end{tabular}
\label{tab:stat1}  
\end{table*}
\linespread{1}

\begin{figure*}[!t]
\includegraphics[width = \linewidth]{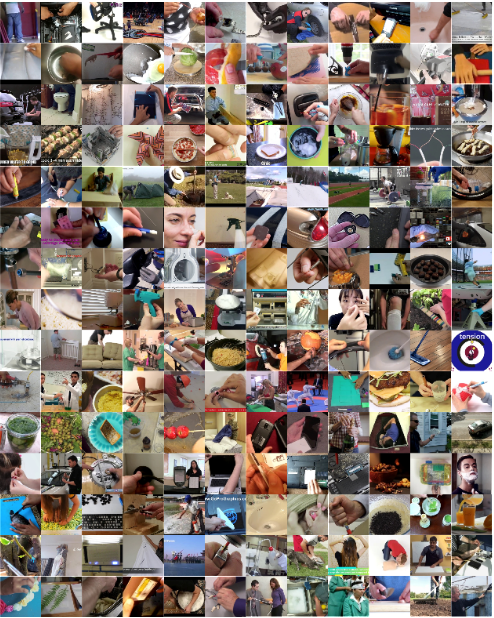}
\caption{An overview of the 180 tasks of the COIN dataset, which are associated to 12 domains to our daily life:
(1) nursing \& caring, (2) vehicles, (3) leisure \& performance, (4) gadgets, (5) electric appliances, (6) household items, (7) science \& craft, (8) plants \& fruits, (9) snacks \& drinks, (10) dishes, (11) sports, and (12) housework.
}
\label{fig:all}
\end{figure*}

\end{appendices}

{
\bibliographystyle{ieee}
\bibliography{egbib}
}

\begin{IEEEbiography}[{\includegraphics[width=1in,height=1.25in,clip,keepaspectratio]{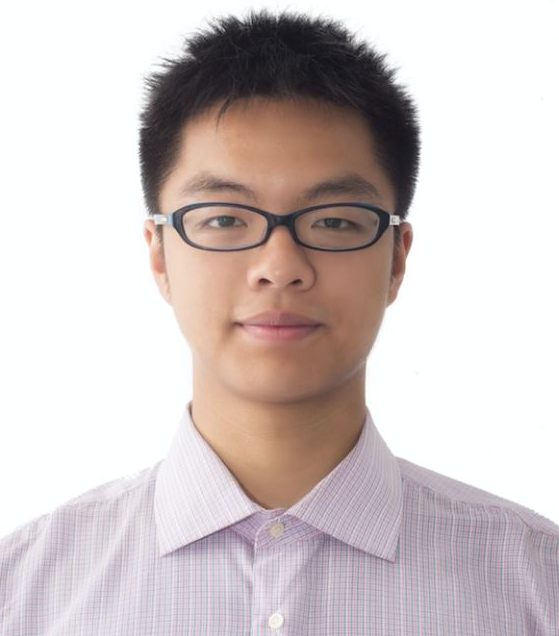}}]{Yansong Tang}
received the B.S. degree in the Department of Automation, Tsinghua University, China, in 2015. He is currently a Ph.D Candidate with the Department of Automation, Tsinghua University, China. 
His current research interest lies in human behaviour understanding for computer vision. 
He has authored 10 scientific papers in this area, where 4 papers are published in top journals and conferences including IEEE TIP, CVPR and ACM MM. 
He serves as a regular reviewer member for a number of journals and conferences, \textit{e.g.,} TPAMI, TIP, TCSVT, CVPR, ICCV, AAAI and so on. He has obtained the National Scholarship of Tsinghua in 2018.
\end{IEEEbiography}

\begin{IEEEbiography}[{\includegraphics[width=1in,height=1.25in,clip,keepaspectratio]{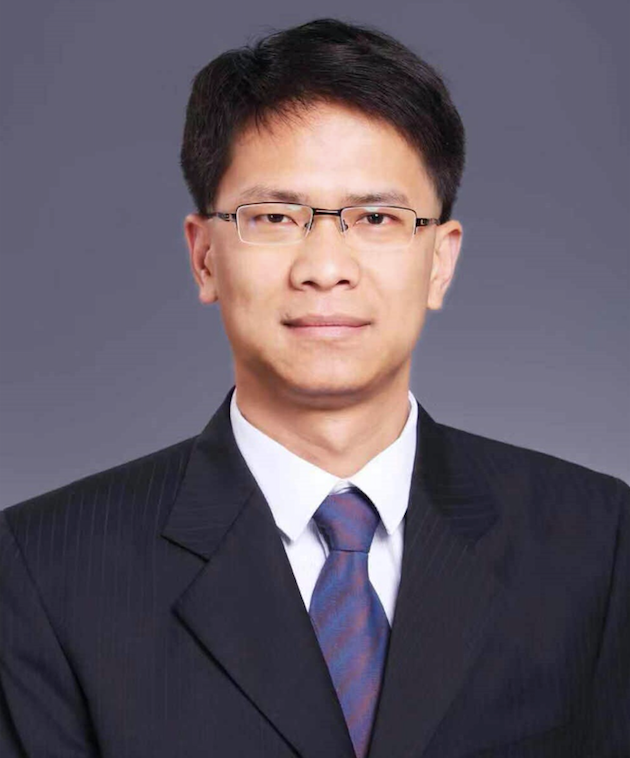}}]{Jiwen Lu}
(M'11-SM'15) received the B.Eng. degree in mechanical engineering and the M.Eng. degree in electrical engineering from the Xi'an University of Technology, Xi'an, China, in 2003 and 2006, respectively, and the Ph.D. degree in electrical engineering from Nanyang Technological University, Singapore, in 2012. He is currently an Associate Professor with the Department of Automation, Tsinghua University, Beijing, China. His current research interests include computer vision, pattern recognition, and machine learning. He has authored/co-authored over 200 scientific papers in these areas, where 70+ of them are IEEE Transactions papers and 50+ of them are CVPR/ICCV/ECCV papers. He serves the Co-Editor-of-Chief of the Pattern Recognition Letters, an Associate Editor of the IEEE Transactions on Image Processing, the IEEE Transactions on Circuits and Systems for Video Technology, the IEEE Transactions on Biometrics, Behavior, and Identity Science, and Pattern Recognition. He is a member of the Multimedia Signal Processing Technical Committee and the Information Forensics and Security Technical Committee of the IEEE Signal Processing Society, and a member of the Multimedia Systems and Applications Technical Committee and the Visual Signal Processing and Communications Technical Committee of the IEEE Circuits and Systems Society. He was a recipient of the National 1000 Young Talents Program of China in 2015, and the National Science Fund of China for Excellent Young Scholars in 2018, respectively. He is a senior member of the IEEE.
\end{IEEEbiography}

\begin{IEEEbiography}[{\includegraphics[width=1in,height=1.25in,clip,keepaspectratio]{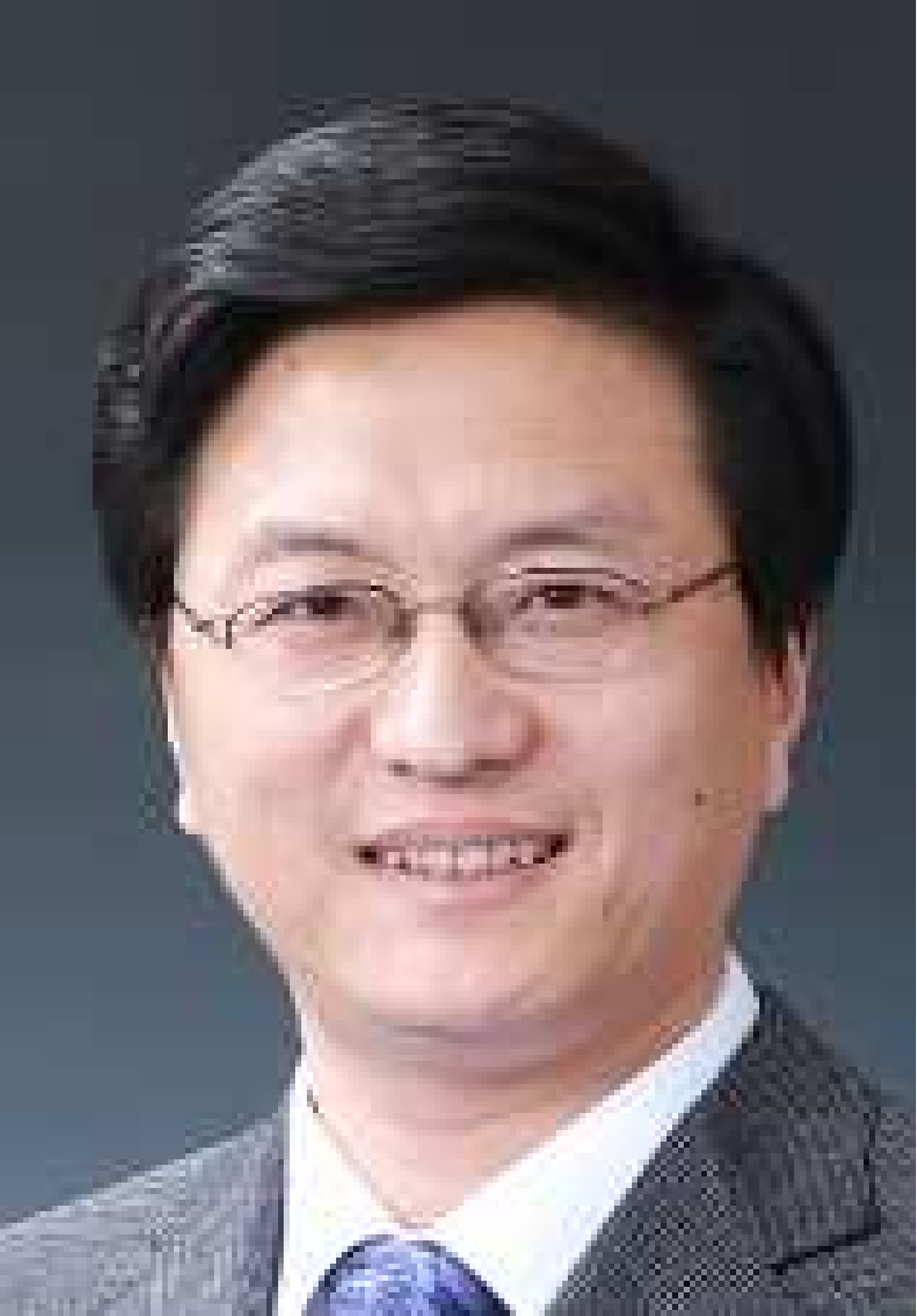}}]{Jie Zhou}
(M'01-SM'04) received the BS and MS degrees both from the Department of Mathematics, Nankai University, Tianjin, China, in 1990 and 1992, respectively, and the PhD degree from the Institute of Pattern Recognition and Artificial Intelligence, Huazhong University of Science and Technology (HUST), Wuhan, China, in 1995. From then to 1997, he served as a postdoctoral fellow in the Department of Automation, Tsinghua University, Beijing, China. Since 2003, he has been a full professor in the Department of Automation, Tsinghua University. His research interests include computer vision, pattern recognition, and image processing. In recent years, he has authored more than 300 papers in peer-reviewed journals and conferences. Among them, more than 60 papers have been published in top journals and conferences such as the IEEE Transactions on Pattern Analysis and Machine Intelligence, IEEE Transactions on Image Processing, and CVPR. He is an associate editor for the IEEE Transactions on Pattern Analysis and Machine Intelligence and two other journals. He received the National Outstanding Youth Foundation of China Award. He is a senior member of the IEEE.
\end{IEEEbiography}

\end{document}